\documentclass[letterpaper,journal]{IEEEtran}
\usepackage{amsmath,amsfonts}
\usepackage{algorithm}
\usepackage{array}
\usepackage{subfigure}
\usepackage{textcomp}
\usepackage{stfloats}
\usepackage{url}
\usepackage{verbatim}
\usepackage{graphicx}
\usepackage{cite}
\usepackage{amssymb}
\usepackage{booktabs}
\usepackage{colortbl}
\usepackage{mathtools}
\usepackage{microtype}
\usepackage{float}
\usepackage{bm}
\usepackage{amsthm}
\usepackage{multirow}
\usepackage{adjustbox}
\usepackage{caption}
\usepackage{subcaption}
\usepackage{algpseudocode}
\usepackage{threeparttable}
\usepackage{pifont}
\usepackage{enumitem}
\usepackage{bbding}
\usepackage{fontawesome}
\usepackage{algorithmicx}
\usepackage{makecell}
\usepackage{rotating}

\usepackage{tabularx}
\usepackage[table]{xcolor}

\usepackage{siunitx}
\definecolor{lightgray}{rgb}{0.9, 0.9, 0.9}
\definecolor{lightblue}{rgb}{0.8, 0.9, 1.0}
\definecolor{lightgreen}{rgb}{0.9, 1.0, 0.9}
\definecolor{lavender}{rgb}{0.9, 0.9, 0.98}
\definecolor{brightgreen}{RGB}{102, 255, 0}
\definecolor{deepgreen}{RGB}{50, 186, 70}

\begin{document}

\title{NAP-Tuning: Neural Augmented Prompt Tuning for Adversarially Robust Vision-Language Models}

\author{Jiaming Zhang, Xin Wang, Xingjun Ma, Lingyu Qiu, Yu-Gang Jiang,~\IEEEmembership{Fellow, ~IEEE}, and Jitao Sang
\thanks{This work was supported in part by the National Natural Science Foundation of China (No. 62576030, 62572040). \emph{(Corresponding authors: Xingjun Ma; Jitao Sang)}}
\thanks{Jiaming Zhang is with the School of Computer Science and Technology, Beijing Jiaotong University, Beijing, China, and also with the Shanghai Key Lab of Intell. Info. Processing, School of CS, Fudan University, Shanghai, China (e-mail: lanzhang1107@gmail.com).}%
\thanks{Xin Wang, Xingjun Ma, Yu-Gang Jiang are with the Shanghai Key Lab of Intell. Info. Processing, School of CS, Fudan University, Shanghai, China (e-mail: xinwang22@m.fudan.edu.cn, xingjunma@fudan.edu.cn, ygj@fudan.edu.cn). }%
\thanks{Lingyu Qiu is with the Department of Mathematics and Applications, University of Naples Federico II, Naples, Italy (e-mail: qiulingyu0925@gmail.com).}
\thanks{Jitao Sang is with the State Key Laboratory of Advanced Rail Autonomous Operation, Beijing Jiaotong University, Beijing, China, and also with the School of Computer Science and Technology, Beijing Jiaotong University, Beijing, China (e-mail: jtsang@bjtu.edu.cn). }
}

\maketitle

\begin{abstract}
Vision-Language Models (VLMs) such as CLIP have demonstrated remarkable capabilities in understanding relationships between visual and textual data through joint embedding spaces. Despite their effectiveness, these models remain vulnerable to adversarial attacks, particularly in the image modality, posing significant security concerns. Building upon our previous work on Adversarial Prompt Tuning (AdvPT), which introduced learnable text prompts to enhance adversarial robustness in VLMs without extensive parameter training, we present a significant extension by introducing the Neural Augmentor framework for Multi-modal Adversarial Prompt Tuning (\emph{NAP-Tuning}).
As a significant extension, NAP-Tuning first establishes a comprehensive multi-modal (text and visual) and multi-layer prompting framework. 
The core of this framework is a targeted structural augmentation for feature-level purification, implemented through our Neural Augmentor approach.
This framework implements feature purification by incorporating TokenRefiners—lightweight neural modules that learn to reconstruct purified features via residual connections—to directly address distortions in the feature space. 
This structural intervention is what enables the multi-modal and multi-layer system to effectively perform modality-specific and layer-specific feature rectification.
Comprehensive experiments demonstrate that NAP-Tuning significantly outperforms existing methods across various datasets and attack types. Notably, our approach shows significant improvements over the strongest baselines under the challenging AutoAttack benchmark, outperforming them by 32.3\% on ViT-B16 and 31.3\% on ViT-B32 architectures while maintaining competitive clean accuracy. This work highlights the efficacy of internal feature-level intervention in prompt tuning for adversarial robustness, moving beyond input-side alignment approaches to create an adaptive defense mechanism that can identify and rectify adversarial perturbations across embedding spaces.
\end{abstract}

\begin{IEEEkeywords}
Adversarial robustness, Vision-Language models, Prompt tuning, Feature purification, Neural augmentation
\end{IEEEkeywords}

\section{Introduction}
Large-scale pre-trained Vision-Language Models (VLMs) have demonstrated remarkable capabilities in understanding and connecting visual and textual information, enabling impressive performance on a wide range of downstream tasks. Models like CLIP \cite{radford2021learning} and ALIGN \cite{jia2021scaling} have shown unprecedented zero-shot transfer abilities by learning from web-scale image-text pairs. As these models gain widespread adoption in real-world applications, ensuring their robustness against adversarial attacks becomes increasingly critical.

In our previous work \cite{zhang2024adversarial}, we introduced Adversarial Prompt Tuning (AdvPT), a novel approach that enhances the adversarial robustness of VLMs by aligning text embeddings with adversarial image embeddings through learnable text prompts. This method represented a paradigm shift from traditional adversarial training \cite{madry2017towards} approaches by focusing on prompt-level modifications rather than model parameter retraining, offering significant computational advantages while maintaining effectiveness.

While AdvPT demonstrated the potential of prompt tuning for adversarial defense, it exhibited three critical limitations: (1) its restriction to text modality prompts, (2) its reliance on single-layer prompting, and (3) its emphasis on loss function redesign rather than architectural innovations. Following our initial work, several approaches \cite{zhou2024few, luo2024adversarial} have attempted to address some of these issues, particularly by extending prompting to multiple modalities and layers. However, these works, while incrementally beneficial, share the same fundamental architectural limitations as AdvPT and thus fail to address the core challenge of robust feature representation.

The fundamental limitation across all existing prompt-based approaches is the direct transplantation of prompt tuning techniques from the generalization domain (\emph{e.g.,} CoOp \cite{zhou2022learning} and MaPLe \cite{khattak2023maple}) to adversarial defense without rethinking the underlying architecture. This overlooks a critical insight: adversarial robustness is a fundamentally harder task than standard generalization, as it requires richer \emph{model capacity} \cite{schmidt2018adversarially, zhang2019theoretically}. 
This requirement for increased capacity creates a fundamental trade-off during VLM adaptation. While freezing the backbone preserves pre-trained knowledge, the limited structural capacity of standard prompt tuning, which typically optimizes a small set of input-side vectors, often fails to rectify corrupted features effectively. Conversely, high-capacity full fine-tuning risks catastrophic forgetting of pre-trained representations \cite{wang2023improved}. We aim to bridge this gap by incorporating a modular structural augmentation that provides targeted capacity for feature purification while maintaining the backbone's integrity. A systematic analysis and positioning of these adaptation paradigms are further detailed in Section \ref{sec:2.4}.

\begin{figure*}[t]
\centering
\subfigure[Original AdvPT approach]{
    \includegraphics[width=0.3\linewidth]{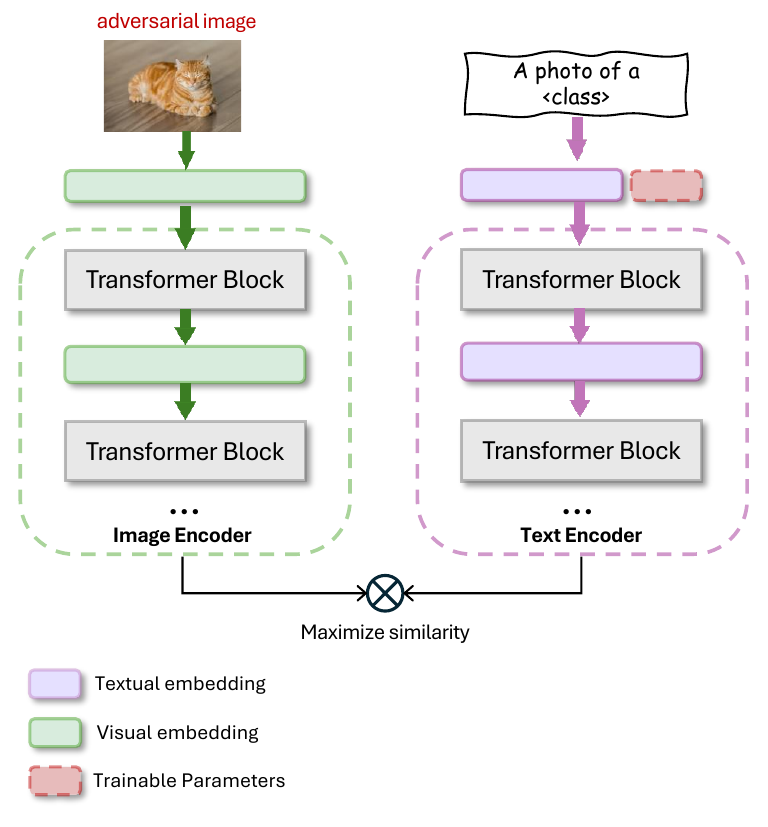}
}
\subfigure[Existing Multimodal AdvPT]{
    \includegraphics[width=0.3\linewidth]{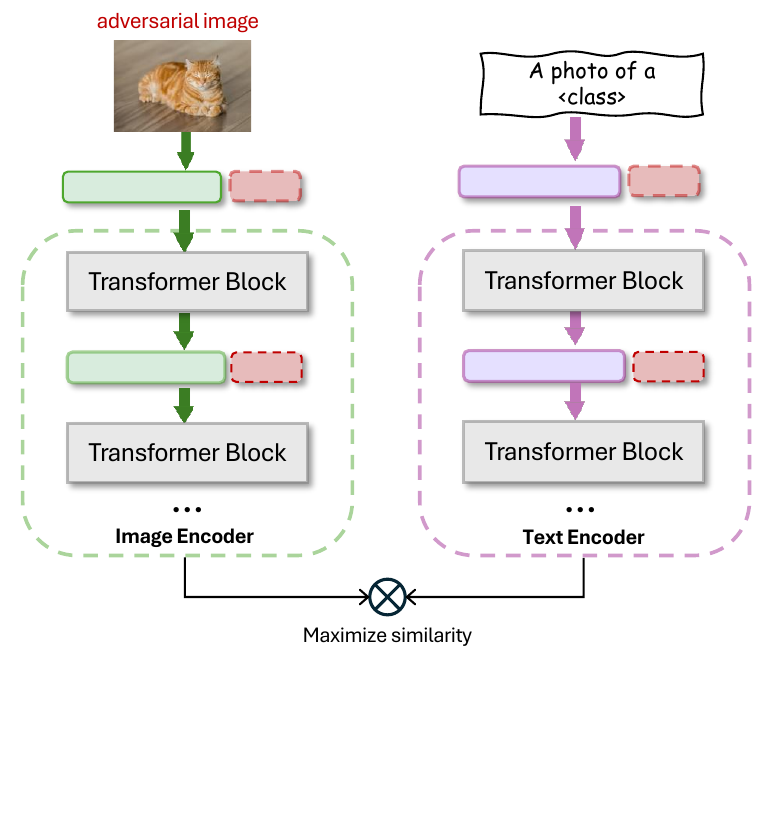}
}
\subfigure[Our proposed NAP-Tuning]{
    \includegraphics[width=0.3\linewidth]{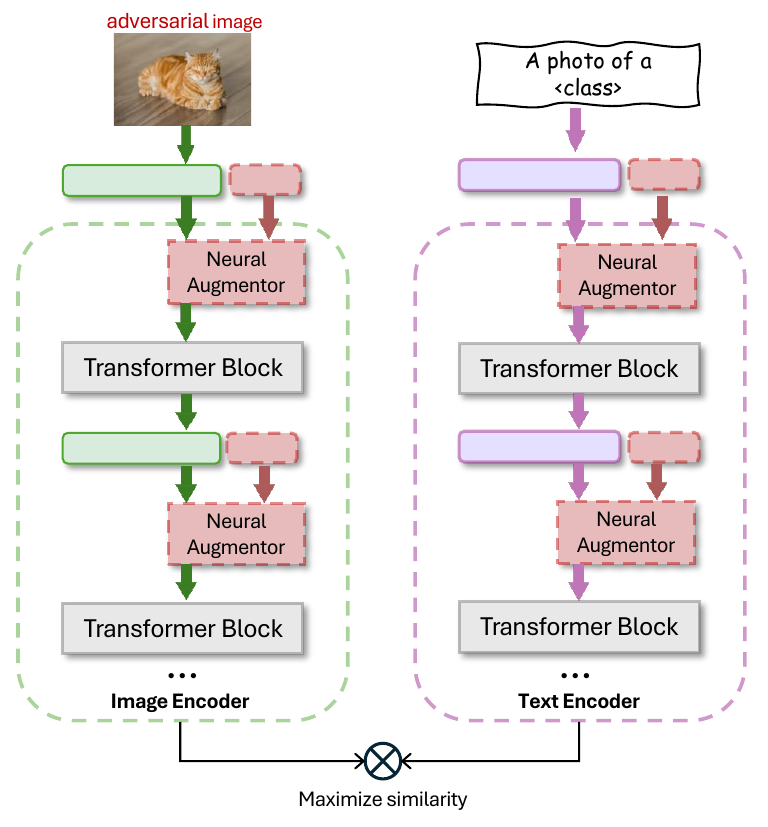}
}
\caption{Comparison of adversarial prompt tuning approaches: (a) The original AdvPT that uses only text prompts, (b) Existing multimodal approaches that incorporate prompts in both visual and text pathways, and (c) Our proposed NAP-Tuning framework that extends multimodal prompting with feature purification via token refiners, enabling the reconstruction of clean feature representations from adversarially perturbed inputs.}
\label{fig:comparison}
\end{figure*}

To address these limitations, we propose the Neural Augmentor framework for Multi-modal Adversarial Prompt Tuning (\textbf{NAP-Tuning}). Our approach introduces a targeted structural augmentation designed to navigate the trade-off between model capacity and knowledge preservation. We keep the backbone frozen to avoid catastrophic forgetting, but introduce a modular neural component: the Neural Augmentor. Unlike standard prompt vectors, the Augmentor is a set of lightweight neural networks (TokenRefiners) that operate directly on the feature representations. This modular design provides the functionality required to model and rectify feature distortions at the token level. Fig. \ref{fig:comparison} illustrates the evolution from text-only prompting to our comprehensive NAP-Tuning.

Through extensive experiments under rigorous evaluation protocols, we demonstrate that our approach significantly outperforms our initial AdvPT and other state-of-the-art methods, achieving superior robustness while maintaining competitive clean accuracy. Our analysis revealed that NAP-Tuning effectively learns to correct adversarial distortions in feature space, providing insights into the working mechanisms of successful adversarial defenses.

As an extension of our conference paper \cite{zhang2024adversarial}, which introduced the concept of \emph{Adversarial Prompt Tuning}, this work makes several significant contributions: 
\begin{itemize}
\item \textbf{Systematic Extension:} Building upon our original text-only AdvPT \cite{zhang2024adversarial}, we implement a coordinated framework across both visual and textual pathways. This systemic extension addresses adversarial vulnerabilities that manifest across modalities, providing a unified multi-layer prompting architecture for robust vision-language adaptation.
\item \textbf{Modular Structural Augmentation:} We introduce the Neural Augmentor, a modular structural augmentation utilizing lightweight TokenRefiners for explicit feature-level purification. This design represents a functional shift from loss-alignment methods to direct feature rectification, providing a plug-and-play mechanism that is easily compatible with various frozen VLM backbones.
\item \textbf{Comprehensive Validation:} We provide extensive experimental validation for our method across 11 datasets and under stricter evaluation settings. Our analysis confirms the effectiveness of the feature purification mechanism, demonstrating state-of-the-art robustness gains while preserving generalization. 
\end{itemize}

\begin{table*}[t]
\centering
\caption{Comparative analysis of Adversarial Adaptation Paradigms for VLMs. We analyze different strategies based on their architectural approach to the robustness-generalization trade-off.}
\label{tab:paradigm_comparison}
\begin{threeparttable}
\begin{tabular}{l c c c c c}
\toprule
\textbf{Adaptation Paradigm (Examples)} & \textbf{Preserves Prior} & \textbf{Increases Model} & \textbf{Architectural} & \textbf{Trainable} & \textbf{Resulting Robustness} \\
& \textbf{Knowledge?} & \textbf{Capacity?} & \textbf{Redesign?} & \textbf{Parameters} & \textbf{Potential} \\
\midrule
(No Adaptation) & \Checkmark & \XSolidBrush & \XSolidBrush & \XSolidBrush & \XSolidBrush \\
\rowcolor{lightgray}
Full Fine-Tuning (TeCoA \cite{mao2023understanding}) & \XSolidBrush & N/A \tnote{a} & \XSolidBrush & High & Low \\
Prompt Tuning (FAP \cite{zhou2024few}, APD \cite{luo2024adversarial}) & \Checkmark & Low & \XSolidBrush & Low & Medium \\
\rowcolor{lightgray}
\textbf{Neural Augmentation (Ours)} & \Checkmark & \textbf{High} & \Checkmark & \textbf{Low-Medium} & \textbf{High} \\
\bottomrule
\end{tabular}
\begin{tablenotes}
    \item[a] This paradigm does not \emph{add} new capacity, but \emph{unlocks} all existing (Very High) parameters for training.
\end{tablenotes}
\end{threeparttable}

\end{table*}

\section{Related Work}
\subsection{Vision-Language Models}
VLMs establish a semantic bridge between visual and textual modalities, typically following two architectural paradigms: generative models based on Large Language Models (e.g., LLaVA \cite{liu2024improved}, Qwen-VL \cite{yang2024qwen2}) and contrastive models (e.g., CLIP \cite{radford2021learning}, ALIGN \cite{jia2021scaling}) that learn joint embedding spaces. While the latter demonstrates exceptional zero-shot transfer capabilities through unified multimodal projections, their inherent susceptibility to adversarial perturbations—particularly in the visual domain—poses significant security risks. Ensuring the robustness of these contrastive embeddings is critical as their deployment expands into safety-critical applications.

\subsection{Prompt Learning}
Originating as a parameter-efficient alternative to full fine-tuning in NLP \cite{liu2021p, li2021prefix}, prompt learning adapts pre-trained models by optimizing input-side transformations. In the VLM context, methods such as CoOp \cite{zhou2022learning} and CoCoOp \cite{zhou2022conditional} replaced manual templates with learnable context vectors. Subsequent innovations, including MaPLe \cite{khattak2023maple} and PromptKD \cite{li2024promptkd}, extended this paradigm to multi-modal pathways. However, as these techniques were primarily engineered for task generalization, their fixed-architecture design presents a fundamental bottleneck when attempting to counteract the structural feature distortions introduced by adversarial attacks.

\subsection{Adversarial Prompt Tuning}
To circumvent the prohibitive computational costs of conventional adversarial training \cite{madry2017towards, zhang2019theoretically, wang2019improving}, AdvPT \cite{zhang2024adversarial} introduced an efficient defense paradigm by aligning text prompts with adversarial image embeddings. This prompted several extensions, such as FAP \cite{zhou2024few}, which focuses on few-shot adversarial consistency, and APD \cite{luo2024adversarial}, which leverages bimodal distillation for robust adaptation. In parallel, test-time optimization strategies (e.g., R-TPT \cite{sheng2025r}, TAPT \cite{wang2024tapt}, TTC \cite{xing2025clip}) have enhanced VLM security during inference. Despite these advancements, existing methods predominantly focus on loss-level alignment of input-side prompt vectors. Our work diverges from these trends by introducing a modular structural augmentation designed specifically for internal feature-level rectification, a mechanism that targets the root cause of feature distortion beyond simple distribution alignment.

\subsection{Architectural Trade-offs in Robust VLM Adaptation}\label{sec:2.4}
The theoretical foundation for adapting VLMs for robustness involves navigating the fundamental trade-off between standard accuracy and adversarial robustness \cite{zhang2019theoretically}. Robust learning theory establishes that achieving adversarial generalization is a more difficult task, requiring significantly higher model capacity than standard generalization \cite{schmidt2018adversarially}. In the context of large pre-trained models, this requirement creates a specific dilemma: while full fine-tuning provides the necessary capacity to learn robust features, it risks the degradation of pre-trained knowledge \cite{wang2023improved, zhang2023imagenet}. Conversely, fixed-architecture prompt tuning preserves generalization but lacks the structural complexity needed to counteract feature-level distortions.

We clarify this landscape in Table \ref{tab:paradigm_comparison}, positioning our work within three dominant adaptation paradigms. Unlike conventional prompt tuning that relies on loss-alignment of input-side vectors, our \textit{Neural Augmentation} paradigm introduces a modular structural intervention. By incorporating internal TokenRefiners, we provide the targeted capacity required for feature purification without compromising the backbone's integrity. This design is fundamentally modular, allowing the rectification mechanism to be integrated into various frozen transformer backbones as a generic plug-in.

\section{Preliminary: Adversarial Prompt Tuning}

To provide context for our extended approach, we first revisit the key concepts of \emph{Adversarial Prompt Tuning} as introduced in our previous work \cite{zhang2024adversarial}.

\subsection{Revisiting CLIP}
We first provide a brief overview of VLMs, with a particular emphasis on the CLIP \cite{radford2021learning} architecture. Although our methods are specifically designed for CLIP, they are potentially extendable to a broader range of VLMs that are based on the contrastive learning paradigm.
CLIP consists of an image encoder $f_I$ and a text encoder $f_T$, which project images and text into a shared embedding space. The model is trained to maximize the similarity between matched image-text pairs while minimizing the similarity between unmatched pairs.

For an image $x$ and a text description $t$, CLIP computes embeddings $f_I(x)$ and $f_T(t)$, and the similarity is calculated as:

\begin{equation}
    s(x, t) = \frac{f_I(x) \cdot f_T(t)}{||f_I(x)|| \cdot ||f_T(t)||}
\end{equation}

For classification tasks, CLIP computes the similarity between an image and a set of text templates describing each class (e.g., ``a photo of a [CLASS]"), and selects the class with the highest similarity.

\subsection{Adversarial Attacks on VLMs}
Adversarial attacks on VLMs can be categorized into two types based on the attacker's access to model components. In our previous work \cite{zhang2024adversarial}, we considered a restricted threat model where the attacker only has access to the image encoder. Under this assumption, adversarial examples are crafted by maximizing the KL divergence between clean and perturbed image features:

\begin{align}
    x_{adv} &= \arg\max_{x'} D_{KL}(f_I(x) , f_I(x')) \nonumber \\
    &\text{s.t.} \quad ||x' - x||_p \leq \epsilon, 
    \label{eq:kl_attack}
\end{align}
where $D_{KL}$ is the KL divergence measure. In this work, we consider a more challenging threat model that assumes the attacker has access to both image and text encoders, including textual prompt. This stronger adversary can generate more effective attacks by maximizing the cross-entropy loss between image-text pairs (even when the text is represented as learnable vectors):

\begin{align}
x_{adv} &= \arg\max_{x'} \mathcal{L}_{CE}( cos(f_I(x'), f_T(\mathbf{T})), y ) \nonumber \\
&\text{s.t.} \quad ||x' - x||_p \leq \epsilon ,
\label{eq:ce_attack}
\end{align}
where $\mathbf{T} = \{t_1, ..., t_k\}$ is the set of all $k$ class text templates, $y$ is the ground-truth class label, $cos(\cdot, \cdot)$ is the cosine similarity function which produces the logits, and $\mathcal{L}_{CE}$ is the standard cross-entropy loss.

\subsection{Basic Adversarial Prompt Tuning}
AdvPT addresses the vulnerability of VLMs to adversarial attacks by learning to align text embeddings with adversarial image embeddings through prompt tuning. The key insight is that this alignment can enhance robustness without requiring modifications to the model architecture. In AdvPT, the standard text template ``a photo of a [CLASS]" is replaced with a template containing learnable context vectors:

\begin{equation}
    (\mathcal{V}_t, y) = [\mathcal{V}_t, \text{[CLASS]}],
\end{equation}
where $\mathcal{V}_t$ are learnable vectors that are optimized to align with adversarial image embeddings, and $y$ represents class label. The training objective for AdvPT is to maximize the similarity between the clean text embedding (with learned prompt) and the adversarial image embedding:
\begin{equation}
\label{eq5}
    \min_{\mathcal{V}_t} \mathcal{L}(x_{adv}, y) = \max_{\mathcal{V}_t} s(x_{adv}, (\mathcal{V}_t, y)),
\end{equation}
where $x_{adv}$ is the adversarial version of image $x$, and $(\mathcal{V}_t, y)$ is the text template for the true class $y$.

\section{Neural Augmentor for Multi-Modal Adversarial Prompt Tuning}

Building upon our preliminary work on AdvPT, we introduce the NAP-Tuning framework as a structural augmentation for prompt-based adversarial defenses. This section details our approach, its structural augmentations, and theoretical foundations.

\subsection{Framework Overview}

\begin{figure}[t]
\centering
\includegraphics[width=0.8\linewidth]{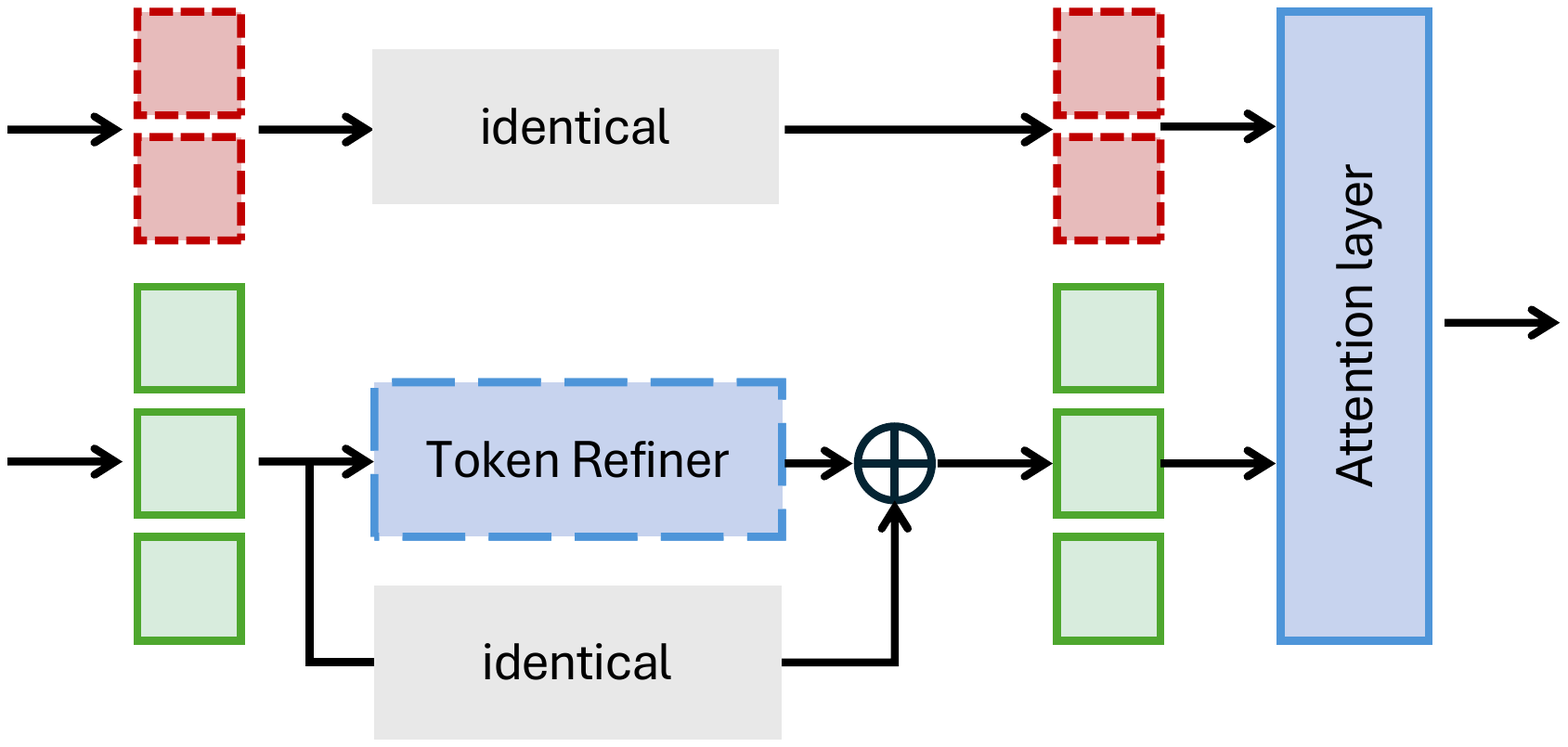}
\caption{Overview of our proposed Neural Augmentor module for multi-modal AdvPT.}
\label{fig:module}
\end{figure}

Fig. \ref{fig:comparison} (c) and Fig. \ref{fig:module} illustrate the NAP-Tuning framework and its core Neural Augmentor module. Conceptually, our approach represents a strategic evolution of the efficiency-oriented paradigm established by AdvPT \cite{zhang2024adversarial}. While the original AdvPT focused on minimizing computational overhead by avoiding backbone backpropagation, NAP-Tuning identifies a superior trade-off: by incorporating lightweight structural augmentations for internal feature purification, we achieve a massive leap in robustness while maintaining an efficiency profile that remains orders of magnitude superior to traditional full fine-tuning. This conceptual shift—moving from purely input-side alignment to internal feature rectification—is implemented through three coordinated components: (1) a multi-modal prompting system for visual-textual coordination, (2) a multi-layer prompt architecture for hierarchical intervention, and (3) Neural Augmentor modules that perform explicit token-level purification via residual refiners.

\subsection{Structural Innovations}

\subsubsection{Multi-Modal Prompting}

A key limitation of our original AdvPT approach was its exclusive focus on the text modality, which left the visual pathway vulnerable to direct attacks. Our enhanced framework addresses this limitation by implementing learnable prompts in both textual and visual pathways, creating a coordinated defense system.

For the text pathway, we build upon the original AdvPT framework with enhanced prompt vectors $(\mathcal{V}_t, y)$. For the visual pathway, we introduce a parallel set of prompt vectors that operate on the visual inputs $(\mathcal{V}_i, x_{adv})$, where $\mathcal{V}_i$ are learnable visual prompt vectors. These visual prompts serve a complementary function to the text prompts, helping to guide perturbed visual representations back toward their clean manifold. Equation \eqref{eq5} can therefore be reformulated as:
\begin{equation}
\label{eq6}
    \min_{\mathcal{V}_i, \mathcal{V}_t} \mathcal{L}(x_{adv}, y) = \max_{\mathcal{V}_i, \mathcal{V}_t} \ s((\mathcal{V}_i, x_{adv}), (\mathcal{V}_t, y)).
\end{equation}
The dual-modal prompt system enables a more comprehensive defense, as it can address attacks that target either modality or the cross-modal alignment. This is particularly important in VLMs, where adversarial perturbations can disrupt the crucial cross-modal matching that underlies the model's performance.

\subsubsection{Multi-Layer Prompt Architecture}

Adversarial perturbations manifest differently across network depths, affecting both low-level perceptual features and high-level semantic representations. Our multi-layer prompt architecture addresses this by placing learnable prompts at multiple depths within the transformer network:
\begin{equation}
\label{layer}
    h^{(l)} = \text{Layer}^{(l)}([\mathcal{V}^{(l)}, h^{(l-1)}]),
\end{equation}
where $h^l$ represents the output of layer $l$, $\mathcal{V}^{(l)}$ are layer-specific learnable prompt vectors, and $\text{Layer}^{(l)}$ is the transformer layer function. This hierarchical prompting structure allows for targeted interventions at different levels of feature abstraction. Let $\mathbf{V}_i = \{\mathcal{V}_i^j\}_{j=1}^l$ and $\mathbf{V}_t = \{\mathcal{V}_t^j\}_{j=1}^l$, then Equation \eqref{eq6} can be reformulated as:
\begin{equation}
\label{eq7}
    \min_{\mathbf{V}_i, \mathbf{V}_t} \mathcal{L}(x_{adv}, y) = \max_{\mathbf{V}_i, \mathbf{V}_t} s((\mathbf{V}_i, x_{adv}), (\mathbf{V}_t, y)).
\end{equation}
The multi-layer design offers several advantages over the single-layer approach of our original AdvPT. It enables depth-specific defense, where different layers can address distinct manifestations of adversarial perturbations, from low-level texture disruptions to high-level semantic shifts. It also facilitates progressive feature refinement, where corrections applied at earlier layers can be further refined by later layers, creating a cascading purification effect. More importantly, the increased parameter capacity allows the model to learn more complex defense strategies without modifying the underlying VLM weights.

This hierarchical structure represents a significant advancement over our preliminary work, which was limited to modifying only the textual prompts. By extending the defense mechanism throughout the network, we create a more robust barrier against adversarial perturbations that might otherwise bypass single-layer defenses.

\subsection{Neural Augmentor Design}

The most significant innovation in our framework is the Neural Augmentor—a specialized neural module designed explicitly for feature purification. Unlike conventional prompt tuning that focuses on optimizing context vectors, our Neural Augmentor actively transforms feature representations to counteract adversarial distortions.

\subsubsection{Feature Distortion Theory}

To formalize our approach, we first characterize the effect of adversarial perturbations in feature space. Let $\mathcal{X}$ denote the input space and $\mathcal{F}$ the feature space. For a clean image $x \in \mathcal{X}$ and its adversarial counterpart $x_{adv} = x + \delta$ , the feature distortion can be expressed as:
\begin{equation}
\label{eq:delta}
    \Delta f_I = f_I(x_{adv}) - f_I(x),
\end{equation}
where $f_I(x)$ represents the feature representation of input $x$. Traditional adversarial defenses aim to make classification boundaries robust to these distortions. In contrast, our Neural Augmentor directly targets the distortion itself, attempting to recover:
\begin{equation}
    \hat{f}_I(x) \approx f_I(x_{adv}) + \Phi(f_I(x_{adv})),
\end{equation}
where $\Phi$ is a learned correction function that approximates $-\Delta f$, effectively ``purifying" the adversarial features (including both intermediate and final feature representations) back toward their clean counterparts.

\subsubsection{TokenRefiner Architecture}

The core component of the Neural Augmentor is the TokenRefiner $R$—a lightweight neural network that processes individual token representations to identify and correct adversarial distortions. For each token representation $z \in \mathbb{R}^d$, the TokenRefiner computes a corrective term $\tilde{z} = R(z)$. The TokenRefiner function is implemented as a two-layer network with a residual connection. It ensures stable gradient flow during training, facilitating effective optimization. Moreover, when no correction is needed (e.g., for clean inputs), the network can learn to output values near zero, effectively preserving the original features through an identity fallback mechanism. The $R$ operates on both text and visual tokens, with modality-specific parameters that allow it to learn distinct correction patterns for each pathway.

We apply the TokenRefiner $R$ to correct potentially perturbed feature representations before combining them with learnable vectors for the attention mechanism. Therefore, Equation \eqref{layer} can be reformulated as:
\begin{equation}
\label{layer_refined}
    h^{(l)} = \text{Layer}^{(l)}\big(\text{Attention}(\mathcal{V}^{(l)}, \tilde{h}^{(l-1)})\big),
\end{equation}
where $\tilde{h}^{(l-1)} = R({h}^{(l-1)})$. Through this attention mechanism, we expect the learnable vectors $\mathcal{V}^{(l)}$ and the augmented feature representations $\tilde{h}^{(l-1)}$ to cooperatively optimize and enhance robustness against adversarial attacks.
Consequently, the final learning objective can be expressed as an extension of Equation \eqref{eq7}:
\begin{equation}
\label{eq8}
\begin{aligned}
    \min_{\substack{\mathbf{V}_i,\, \mathbf{V}_t, \\ \theta_i,\, \theta_t}} \mathcal{L}(x_{adv}, y) 
    &= \max_{\substack{\mathbf{V}_i,\, \mathbf{V}_t, \\ \theta_i,\, \theta_t}} 
    s((\mathbf{V}_i, x_{adv}; R_{i}), (\mathbf{V}_t, y; R_{t})),
\end{aligned}
\end{equation}
where $\theta_i$ and $\theta_t$ represent the learnable parameters of the TokenRefiner $R$ in the visual and text branches, respectively.

The Neural Augmentor implements a hierarchical purification mechanism across token, layer, and modality dimensions, enabling localized and structural rectification of adversarial artifacts. This approach represents a conceptual shift from treating robustness as a classification boundary problem (typical of input-side alignment methods) to directly targeting the root cause: the distortion of internal feature representations. By explicitly intercepting and restoring perturbed tokens within the backbone's feature flow, our design maintains the model's representational integrity and cross-modal alignment even under sophisticated adversarial threats.

\subsection{Training Methodology}
The efficacy of our NAP-Tuning framework relies on a theoretically grounded training methodology that addresses the fundamental challenge in adversarial learning: optimizing the trade-off between standard accuracy and adversarial robustness while maintaining feature-space integrity. Unlike conventional adversarial training approaches that focus primarily on decision boundary robustification, our method directly targets feature-level purification—a strategy particularly well-aligned with the core operating principle of VLMs, which fundamentally depends on cross-modal feature alignment.

\subsubsection{Adversarial Example Generation}
During training, we generate adversarial examples using the Projected Gradient Descent (PGD) method with the following objective:
\begin{equation}
    x_{adv} = \arg\max_{x'} \mathcal{L}_{adv}(f_\theta(x'), y) \quad \text{s.t.} \quad \|x' - x\|_\infty \leq \epsilon,
\end{equation}
where $f_\theta$ represents our model and $\mathcal{L}_{adv}$ denotes the cross-entropy loss. We employ a standard multi-step PGD implementation to generate strong adversarial examples, ensuring that our defense is trained against sophisticated perturbations.

Crucially, the feature-level corrections learned by our model exhibit transferability across different attack types due to the commonality in how various attacks distort the underlying feature manifold. This transferability represents a significant advantage of our feature purification approach over methods that merely harden decision boundaries against specific attack patterns.

\subsubsection{Principled Loss Formulation}
In contrast to previous defense methods that employ complex, multi-term loss functions with heuristically determined components, we derive a principled, minimal training objective from the theoretical foundations of robust learning:
\begin{equation}
    \mathcal{L}(\theta) = \mathcal{L}_{clean}(\theta) + \alpha(\tau) \cdot \mathcal{L}_{adv}(\theta),
\end{equation}
where $\alpha(\tau)$ is a theoretically motivated dynamic balancing coefficient that evolves with training epoch $\tau$. The clean loss $\mathcal{L}_{clean}$ and adversarial loss $\mathcal{L}_{adv}$ are defined as:
\begin{align}
    \mathcal{L}_{clean}(\theta) &= \mathbb{E}_{(x,y) \sim \mathcal{D}}[\mathcal{L}(f_\theta(x), y)], \\
    \mathcal{L}_{adv}(\theta) &= \mathbb{E}_{(x,y) \sim \mathcal{D}}[\mathcal{L}(f_\theta(x_{adv}), y)],
\end{align}
where $\mathcal{D}$ represents the data distribution, and expectations are taken over the training dataset.

The dynamic balancing coefficient $\alpha(\tau)$ follows a sigmoid schedule that systematically transitions from emphasizing clean performance to prioritizing adversarial robustness:
\begin{equation}
\label{alpha}
\alpha(\tau) = \alpha_0 \cdot \frac{1}{1 + \exp\left( -10 \left( \frac{\tau}{T_{max}} - 0.5 \right) \right)},
\end{equation}
where $\alpha_0$ is the maximum weight assigned to the adversarial loss, and $T_{max}$ denotes the total number of training epochs. This schedule implements a curriculum learning strategy that allows the model to first establish representational capacity on clean data before gradually adapting to adversarial inputs.

The theoretical justification for this scheduling approach stems from optimization landscape analysis: the loss surface for adversarial examples typically contains sharper curvature and more local minima than that of clean examples. By initially focusing on clean examples, we guide optimization toward regions of the parameter space with favorable generalization properties before refining these parameters to accommodate adversarial inputs. This procedure effectively navigates the complex optimization landscape of robust learning while mitigating the well-documented trade-off between standard accuracy and adversarial robustness.

\begin{table*}[!t]
  \centering
\caption{Evaluation results on ViT-B16 under clean and adversarial settings. Performance is reported across multiple datasets under white-box (PGD and AutoAttack) and black-box attacks (M-Attack), where $\dagger$ presents our method. Best results are highlighted in \textbf{bold}.}
\label{tab:vitb16}
  \sisetup{detect-weight=true, detect-inline-weight=math}
  \setlength{\tabcolsep}{3pt} 
  \small 
  \renewcommand{\arraystretch}{0.8}
  \begin{tabular*}{\textwidth}{@{\extracolsep{\fill}} c l *{12}{S[table-format=2.1]} @{}}
  	\toprule

      & 
      & \rotatebox{0}{{\textbf{ImageNet}}} & \rotatebox{0}{{\textbf{Caltech}}} & \rotatebox{0}{{\textbf{DTD}}}
      & \rotatebox{0}{{\textbf{Eurosat}}} & \rotatebox{0}{{\textbf{Aircraft}}} & \rotatebox{0}{{\textbf{Food}}}
      & \rotatebox{0}{{\textbf{Flowers}}} & \rotatebox{0}{{\textbf{Pets}}} & \rotatebox{0}{{\textbf{Cars}}}
      & \rotatebox{0}{{\textbf{SUN}}} & \rotatebox{0}{{\textbf{UCF}}} & \rotatebox{0}{{\textbf{Avg}}} \\
    \midrule
    \multirow{8}{*}{\rotatebox{90}{\textbf{Clean}}} 
      & Vanilla        & 66.7  & 93.3  & 44.1  & 48.3  & 24.7 & {\bfseries 85.8} & 70.7  & 89.1  & 65.6 & 62.6  & 67.5  & 65.3  \\
      & FAP   & 60.1  & 92.8  & 60.7  & 68.3  & 26.0 & 72.5 & 84.6  & 87.7  & 60.9 & 65.2  & 75.1  & 68.5  \\
      & APD   & 57.7  & 90.2  & 57.3  & 65.7  & 28.9 & 78.2 & 85.7  & 84.7  & 58.8 &   59.6  & 75.8  &  67.5     \\
      & AdvPT          & {\bfseries 67.5} & {\bfseries 94.1} & {\bfseries 69.6} & 69.0 & 28.7 & 85.0           & 87.9  & {\bfseries 91.7} & 70.8 & {\bfseries 70.7} & {\bfseries 77.4} & {\bfseries 73.9} \\  
      & AdvPT-V        & 60.7  & 89.8  & 36.9  & 28.1  & 16.1 & 63.8           & 54.4  & 84.7  & 51.7 & 58.4  & 54.1  & 54.4  \\  
      & AdvPT-VLI      & 63.8  & 85.7  & 37.2  & 18.2  & 12.0 & 62.3           & 53.6  & 78.7  & 46.9 & 52.8  & 50.5  & 51.1  \\  
      & AdvPT-VLJ      & 65.9  & 88.0  & 37.8  & 20.0  & 10.3 & 69.6           & 59.7  & 82.1  & 53.5 & 56.6  & 55.8  & 54.5  \\  
      & NAP-Tuning\rlap{\textsuperscript{$\dagger$}}
                       & 57.6  & 92.4  & 59.7  & {\bfseries 83.0} & {\bfseries 56.3} & 60.6           & {\bfseries 96.1} & 82.4  & {\bfseries 84.6} & 62.2  & 73.4  & 73.5  \\  
    \midrule[\heavyrulewidth]
     \multirow{8}{*}{\rotatebox{90}{\textbf{M-Attack}}}  
       & Vanilla        & 21.3  & 37.3  & 14.6  & 16.8  & 13.9 & 35.2           & 21.5  & 38.6  & 20.1 & 33.0  & 26.9  & 25.4  \\  
       & FAP            & 43.6  & 70.5  & 30.1  & 11.6  & 24.0 & 48.2           & 42.5  & 55.7  & 42.6 & 45.5  & 36.8  & 41.0  \\
       & APD            & 41.9  & 71.6  & 25.4  & 15.7  & 19.0 & 42.3           & 36.4  & 57.0  & 41.4 & 40.3  & 33.3  & 38.6  \\  
       & AdvPT          & 34.4  & 43.6  & 12.8  & 8.8   & 9.6  & 38.5           & 29.0  & 47.5  & 32.6 & 39.0  & 45.8  & 31.1  \\  
       & AdvPT-V        & 45.3  & 67.3  & 25.7  & 13.6  & 17.8 & 46.3           & 39.5  & 56.4  & 44.4 & 39.7  & 37.4  & 39.4  \\
       & AdvPT-VLI      & 47.9  & 66.4  & 31.5  & 14.8  & 16.1 & 46.7           & 40.6  & 53.2  & 49.7 & 48.0  & 45.7  & 41.9  \\  
       & AdvPT-VLJ      & 40.6  & 71.7  & 24.8  & 18.9  & 19.5 & 45.4           & 40.2  & 51.9  & 48.5 & 50.2  & 54.7  & 42.4  \\
       & NAP-Tuning\rlap{\textsuperscript{$\dagger$}}
                        & {\bfseries 49.8}  & {\bfseries 81.2} & {\bfseries 42.9} & {\bfseries 57.3} & {\bfseries 38.7} & {\bfseries 50.0}           & {\bfseries 76.5} & {\bfseries 62.6}  & {\bfseries 71.3} & {\bfseries 57.1}  & {\bfseries 59.9} & {\bfseries 58.8} \\
    \midrule[\heavyrulewidth]
     \multirow{8}{*}{\rotatebox{90}{\textbf{PGD}}} %
      & Vanilla        & 2.7   & 28.3  & 2.0   & 0.1   & 0.0  & 14.4           & 3.2   & 8.8   & 1.4  & 1.4   & 4.0   & 6.0   \\  
      & FAP            & 22.3  & 67.2  & 25.5  & 30.5  & 22.9 & 20.7           & 75.3  & 46.6  & 33.0 & 31.9  & 41.1  & 37.8  \\
      & APD            & 23.6  & 71.4  & 31.5  & 31.0  & 23.9 & 25.2           & 74.3  & 44.9  & 48.7 & 33.3  & 40.0  & 40.7  \\
      & AdvPT          & 1.5   & 27.0  & 6.6   & 0.2   & 0.6  & 0.9            & 4.3   & 2.8   & 0.7  & 1.6   & 2.1   & 4.4   \\  
      & AdvPT-V        & 19.6  & 61.5  & 18.6  & 8.1   & 3.8  & 12.4           & 25.0  & 38.0  & 9.8  & 17.3  & 17.0  & 21.0  \\  
      & AdvPT-VLI      & 21.8  & 59.1  & 18.6  & 10.0  & 2.9  & 12.9           & 21.7  & 38.2  & 10.0 & 15.7  & 14.2  & 20.5  \\  
      & AdvPT-VLJ      & 20.0  & 58.7  & 16.4  & 10.2  & 2.3  & 11.4           & 20.5  & 32.8  & 8.6  & 14.9  & 12.7  & 19.0  \\  
      & NAP-Tuning\rlap{\textsuperscript{$\dagger$}}
                       & {\bfseries 29.6} & {\bfseries 80.8} & {\bfseries 38.9} & {\bfseries 48.7} & {\bfseries 34.4} & {\bfseries 28.8}           & {\bfseries 86.6} & {\bfseries 52.1} & {\bfseries 64.8} & {\bfseries 35.2} & {\bfseries 49.5} & {\bfseries 49.9} \\  
    \midrule[\heavyrulewidth]
    \multirow{8}{*}{\rotatebox{90}{\textbf{AutoAttack}}} 
      & Vanilla        & 0.0   & 0.0   & 0.1   & 0.1   & 0.1  & 0.0            & 0.0   & 0.0   & 0.1  & 0.0   & 0.1   & 0.0   \\
      & FAP            & 12.2   & 54.8   & 12.9  & 2.2   & 1.9   & 8.6  & 19.4         & 31.3   & 3.7   & 13.5  & 17.0    & 16.1   \\
      & APD            & 10.6   & 48.2   & 14.0  & 3.6   & 2.3   & 5.8  & 13.9         & 27.5   & 4.0   & 9.9  & 10.2    & 13.6   \\
      & AdvPT          & 0.1   & 0.0   & 0.2   & 0.2   & 0.4  & 0.1            & 0.1   & 0.1   & 0.1  & 0.1   & 0.0   & 0.1   \\
      & AdvPT-V        & 14.9  & 55.3  & 15.0  & 2.2   & 2.2  & 8.6            & 18.8  & 32.1  & 5.9  & 12.5  & 13.0  & 16.4  \\  
      & AdvPT-VLI      & 17.0  & 54.3  & 14.3  & 9.2   & 1.6  & 9.1            & 16.2  & 33.6  & 4.8  & 11.4  & 11.3  & 16.6  \\  
      & AdvPT-VLJ      & 10.3  & 43.9  & 10.3  & 3.7   & 1.2  & 4.4            & 9.7   & 19.8  & 1.9  & 6.7   & 6.0   & 10.7  \\  
      & NAP-Tuning\rlap{\textsuperscript{$\dagger$}}
                       & {\bfseries 28.3} & {\bfseries 80.6} & {\bfseries 38.4} & {\bfseries 44.4} & {\bfseries 34.1} & {\bfseries 28.3}           & {\bfseries 86.2} & {\bfseries 50.7} & {\bfseries 64.2} & {\bfseries 33.6} & {\bfseries 48.7} & {\bfseries 48.9} \\  
    \bottomrule
  \end{tabular*}
\end{table*}

\begin{table*}[!t]
  \centering
\caption{Evaluation results on ViT-B32 under clean and adversarial settings. Performance is reported across multiple datasets under white-box (PGD and AutoAttack) and black-box attacks (M-Attack), where $\dagger$ presents our method. Best results are highlighted in \textbf{bold}.}
\label{tab:vitb32}
  \sisetup{detect-weight=true, detect-inline-weight=math}
  \setlength{\tabcolsep}{3pt}
  \small
  \renewcommand{\arraystretch}{0.8}
  \begin{tabular*}{\textwidth}{@{\extracolsep{\fill}} c l *{12}{S[table-format=2.1]} @{}}
    \toprule
      & 
      & \rotatebox{0}{\textbf{ImageNet}}
      & \rotatebox{0}{\textbf{Caltech}}
      & \rotatebox{0}{\textbf{DTD}}
      & \rotatebox{0}{\textbf{Eurosat}}
      & \rotatebox{0}{\textbf{Aircraft}}
      & \rotatebox{0}{\textbf{Food}}
      & \rotatebox{0}{\textbf{Flowers}}
      & \rotatebox{0}{\textbf{Pets}}
      & \rotatebox{0}{\textbf{Cars}}
      & \rotatebox{0}{\textbf{SUN}}
      & \rotatebox{0}{\textbf{UCF}}
      & \rotatebox{0}{\textbf{Avg}} \\
    \midrule
    \multirow{8}{*}{\rotatebox{90}{\textbf{Clean}}}
      & Vanilla\phantom{\textsuperscript{$\dagger$}}
        & 62.0  & 91.4  & 44.2  & 45.4  & 19.3  & {\bfseries 80.4} & 66.6  & 87.4  & 60.2  & 62.1  & 63.5  & 62.0  \\
      & FAP\phantom{\textsuperscript{$\dagger$}}
        & 53.8  & 90.4  & 57.3  & 45.9  & 22.0  & 64.8           & 81.9  & 81.2  & 50.6  & 60.8  & 63.5  & 61.1  \\
      & APD\phantom{\textsuperscript{$\dagger$}}
        & 53.1  & 89.9  & 55.7  & 50.2  & 25.6  & 69.0          	& 85.3  & 83.4  & 49.8  & 60.9  & 63.9  & 62.4  \\
      & AdvPT\phantom{\textsuperscript{$\dagger$}}
        & {\bfseries 63.5} & {\bfseries 93.5} & {\bfseries 64.1} & 79.0 & 27.5 & 79.0           & 91.3  & {\bfseries 88.5} & 70.5  & {\bfseries 71.9} & {\bfseries 77.8} & {\bfseries 73.3} \\
      & AdvPT-V\phantom{\textsuperscript{$\dagger$}}
        & 58.8  & 90.2  & 40.8  & 24.6  & 17.1  & 70.4          	& 59.1  & 83.6  & 50.3  & 59.5  & 57.1  & 55.6  \\
      & AdvPT-VLI\phantom{\textsuperscript{$\dagger$}}
      	& 61.4  & 89.2  & 33.3  & 24.9  & 10.0  & 68.6          	& 54.3  & 83.9  & 47.1  & 54.3  & 56.9  & 53.1  \\
      & AdvPT-VLJ\phantom{\textsuperscript{$\dagger$}}
      	& 61.4  & 87.0  & 31.1  & 19.2  & 9.3  & 65.7          	& 49.5  & 83.6  & 46.0  & 53.9  & 55.2  & 51.1  \\
      & NAP-Tuning\rlap{\textsuperscript{$\dagger$}}
      	& 48.9  & 89.8  & 55.1  & {\bfseries 80.4} & {\bfseries 48.2} & 60.2          	& {\bfseries 94.2} & 78.0  & {\bfseries 77.2} & 59.2  & 69.3  & 69.1  \\
    \midrule[\heavyrulewidth]
    \multirow{8}{*}{\rotatebox{90}{\textbf{M-Attack}}}
      & Vanilla
      	& 18.3  & 33.7  & 15.0  & 11.9  & 13.1  & 31.5          	& 18.7  & 35.2  & 19.5  & 28.4  & 22.6  & 22.5  \\
      & FAP
      	& 39.7 & 65.6 & 28.3 & 9.6 & 21.5 & 48.4 & 39.8  & 51.5 & 48.8  & 40.2 & 32.0  & 38.7  \\  
      & APD
      	& 38.4  & 68.2  & 22.9  & 16.1  & 16.9  & 40.6          	& 33.7  & 53.6  & 39.6  & 36.8  & 30.1  & 36.1  \\  
      & AdvPT
      	& 32.0  & 41.5  & 11.9  & 9.2  & 6.7  & 33.7          	& 25.8  & 43.1  & 29.8  & 27.6 & 42.1  & 27.6 \\  
      & AdvPT-V
      	& 40.9  & 69.3  & 33.6  & 3.6  & 12.3  & 49.3          	& 46.7  & 50.9  & 37.9  & 48.9  & 44.3  & 39.8  \\
      & AdvPT-VLI
      	& 42.6  & 65.8  & 28.3  & 11.9  & 13.1  & 42.7          	& 34.8  & 52.5  & 47.6  & 46.0  & 42.1  & 38.9  \\  
      & AdvPT-VLJ
      	& 42.9  & 66.1  & 28.2  & 14.8  & 9.5  & 45.6          	& 33.9  & 55.8  & 48.1  & 44.7  & 45.3  & 39.5  \\
      & NAP-Tuning
      	& {\bfseries 44.6} & {\bfseries 76.5} & {\bfseries 37.1} & {\bfseries 51.9} & {\bfseries 36.8} & {\bfseries 52.1} & {\bfseries 67.4} & {\bfseries 57.3} & {\bfseries 67.7} & {\bfseries 50.9} & {\bfseries 48.3} & {\bfseries 53.7} \\
    \midrule[\heavyrulewidth]
    \multirow{8}{*}{\rotatebox{90}{\textbf{PGD}}}
      & Vanilla
      	& 2.0   & 29.1  & 4.9   & 0.2   & 0.0  	& 7.4          	& 1.7   & 4.2   & 0.4  	& 1.5  	& 3.0  	& 4.9  	\\  
      & FAP  & 24.5  & 66.8  & 27.1  & 28.3  & 19.2 & 19.8          	& 69.7  & 40.5  & 29.9 & 30.3  & 36.9  & 35.7  \\
      & APD            
      & 18.4  & 66.0  & 30.8  & 22.4  & 20.2 & 26.1          	& 64.5  & 33.9  & 36.8 & 30.4  & 36.8 & 35.1 \\
      & AdvPT
      	& 2.8   & 31.7  & 10.0  & 0.4  	& 0.8  	& 1.8          	& 8.7  	& 3.2  	& 2.0  	& 2.9  	& 4.5  	& 6.3  	\\
      & AdvPT-V
      	& 19.0  & 64.6  & 19.8  & 0.3  	& 2.9  	& 15.0          	& 23.0  & 36.3  & 9.6  	& 18.9  & 18.3  & 20.7  \\  
      & AdvPT-VLI
      	& 20.7  & 62.1  & 16.3  & 5.7  	& 1.5  	& 15.9          	& 19.3  & 37.6  & 8.4  	& 16.2  & 17.7  & 20.1  \\  
      & AdvPT-VLJ
      	& 20.9  & 59.9  & 15.2  & 10.1  & 2.0  	& 15.3          	& 18.8  & 37.5  & 9.6  	& 16.9  & 18.2  & 20.4  \\  
      & NAP-Tuning
      	& {\bfseries 26.5} & {\bfseries 77.9} & {\bfseries 34.2} & {\bfseries 49.1} & {\bfseries 25.4} & {\bfseries 28.4} & {\bfseries 82.4} & {\bfseries 41.3} & {\bfseries 52.4} & {\bfseries 31.5} & {\bfseries 45.0} & {\bfseries 44.9} \\
  	\midrule[\heavyrulewidth]
  	\multirow{8}{*}{\rotatebox{90}{\textbf{AutoAttack}}}
      & Vanilla
      	& 0.0   & 0.4   & 0.3  	& 0.1  	& 0.1  	& 0.0          	& 0.0  	& 0.0  	& 0.0  	& 0.0  	& 0.1  	& 0.1  	\\
      & FAP            & 10.9  & 42.8  & 8.5  & 1.8  	& 1.0  	& 7.1  & 16.7         & 26.2  & 2.4  	& 10.0  & 12.5  & 12.7  \\
      & APD            & 9.0  	& 46.5  & 10.1  & 0.2 & 0.5  	& 6.5  & 14.8         & 17.9  & 1.9  	& 9.4  & 12.2  & 11.7  \\  
      & AdvPT
      	& 0.0  	& 0.6  	& 0.5  	& 0.1  	& 0.2  	& 0.0          	& 0.0  	& 0.1  	& 0.0  	& 0.1  	& 0.0  	& 0.1  	\\  
      & AdvPT-V
      	& 8.3  	& 44.8  & 12.4  & 0.0  	& 0.5  	& 5.4          	& 8.8  	& 16.3  & 2.8  	& 7.4  	& 7.2  	& 10.4  \\  
      & AdvPT-VLI
      	& 9.3  	& 44.2  & 10.5  & 0.2  	& 0.4  	& 5.6          	& 7.6  	& 16.4  & 2.1  	& 6.2  	& 6.8  	& 9.9  	\\  
      & AdvPT-VLJ
      	& 10.1  & 44.7  & 10.3  & 1.2  	& 0.6  	& 6.1          	& 7.9  	& 18.9  & 2.3  	& 7.2 & 8.3  	& 10.7  \\  
      & NAP-Tuning\rlap{\textsuperscript{$\dagger$}}
      	& {\bfseries 20.9} & {\bfseries 77.7} & {\bfseries 34.0} & {\bfseries 48.8} & {\bfseries 25.2} & {\bfseries 28.0} & {\bfseries 82.3} & {\bfseries 40.9} & {\bfseries 50.7} & {\bfseries 31.1} & {\bfseries 44.6} & {\bfseries44.0} \\
    \bottomrule
  \end{tabular*}
\end{table*}

\section{Experiments}\label{sec:exp}

\subsection{Experimental Setup}\label{sec:setup}

\subsubsection{Datasets and Models}
To comprehensively evaluate the effectiveness of our approach, we conduct tests on 11 widely used image classification datasets, including ImageNet \cite{russakovsky2015imagenet}, Caltech101 \cite{FeiFei2004LearningGV}, DTD \cite{cimpoi2014describing}, EuroSAT \cite{helber2019eurosat}, FGVC Aircraft \cite{maji2013fine}, Food101 \cite{bossard2014food}, Oxford Flowers \cite{nilsback2008automated}, Oxford Pets \cite{parkhi2012cats}, Stanford Cars \cite{krause20133d}, SUN397 \cite{xiao2010sun}, and UCF101 \cite{soomro2012ucf101}. These datasets encompass a diverse range of visual tasks from fine-grained recognition to scene classification. We follow the training and testing splits defined in \cite{zhou2022learning}. For the ImageNet test set, consistent with prior adversarial attack studies \cite{dong2018boosting, wang2021enhancing, xie2019improving}, we randomly sample 1,000 images, ensuring one image per class. Our experiments are conducted on the CLIP model. In accordance with previous work, we use hand-crafted prompts as textual inputs (e.g., ``\texttt{a photo of a <class>, a type of pet}'' for Pets).

To evaluate adversarial robustness, we implement both white-box and black-box adversarial attacks. For white-box attacks, we use PGD-100 \cite{madry2017towards} and AutoAttack \cite{croce2020reliable}. Consistent with our training protocol, we use a perturbation magnitude of $\epsilon=1/255$. This provides a direct evaluation of the robustness learned at the training budget. For black-box attacks, we utilize the strong M-Attack \cite{li2025frustratingly}. As black-box attacks are inherently less effective, and to ensure a challenging and meaningful evaluation, we follow standard evaluation practice and use a larger perturbation budget of $\epsilon=8/255$.

\subsubsection{Defense Baselines}
We compare NAP-Tuning against state-of-the-art prompt-based defenses. These include the original AdvPT (text-only) and its multi-modal extensions: AdvPT-V (vision-only), AdvPT-VLI (independent dual-modal), and AdvPT-VLJ (joint mapping akin to MaPLe~\cite{khattak2023maple}). We also include recent strong baselines: FAP~\cite{zhou2024few}, which optimizes a few-shot robustness objective, and APD~\cite{luo2024adversarial}, which utilizes bimodal distillation from a clean teacher.

\subsubsection{Implementation Details}
Our training framework spans 90 epochs with a batch size of 32. We adopt the AdamW optimizer with an initial learning rate of $1\times10^{-4}$, which is modulated through a cosine annealing schedule. For adversarial example generation during training, we implement PGD-5 with a default perturbation magnitude $\epsilon = 1/255$ for training phases, though we analyze the effects of different $\epsilon$ values in subsequent sections.
For the TokenRefiner $R$, we implement a default 2-layer neural network architecture, with the impact of network depth examined in our ablation studies. Regarding the multi-layer prompt architecture, our default configuration utilizes all 12 transformer layers, with layer-specific effects analyzed in detailed experiments. The hyperparameter $\alpha_0$ in Equation \eqref{alpha} is set to 5.0. All experiments are conducted on a single NVIDIA A800 80GB GPU.

\subsection{Main Results and Comparisons} \label{sec:main_results}

We begin by evaluating the overall performance of NAP-Tuning against existing methods. We first compare it to state-of-the-art prompt-based defenses to establish its effectiveness, and then benchmark it against fine-tuning based approaches to highlight the architectural trade-offs.

\subsubsection{Comparison with State-of-the-Art Baselines} 
The main evaluation results are summarized in Tables~\ref{tab:vitb16} and~\ref{tab:vitb32}. Overall, NAP-Tuning achieves highly competitive clean accuracy compared to existing baselines. While its clean performance is slightly lower than that of AdvPT, this is expected, as AdvPT explicitly prioritizes clean accuracy in the robustness-accuracy trade-off, often at the cost of adversarial robustness.

NAP-Tuning demonstrates consistent advantages across scenarios. Under the black-box M-Attack, it shows superior generalization, surpassing the strongest baselines by 16.4\% on ViT-B16 and 13.9\% on ViT-B32. This lead is further amplified under strenuous white-box attacks (e.g., PGD, AutoAttack). Notably, on the rigorous AutoAttack benchmark, our method achieves 48.9\% on ViT-B16 and 44.0\% on ViT-B32 (see Tables~\ref{tab:vitb16} and~\ref{tab:vitb32}), significantly outperforming all competitors.

\subsubsection{Comparison with Fine-tuning Defense Methods} \label{sec:exp_finetuning}

We compare NAP-Tuning with TeCoA~\cite{mao2023understanding}, which fine-tunes CLIP's encoder layers for adversarial robustness. This comparison directly addresses the fundamental question: should we modify the pre-trained encoder or keep it frozen? Both methods are evaluated on ViT-B32 with adversarial examples generated using PGD-100.

\begin{figure}[!t] 
\centering 
\includegraphics[width=\linewidth]{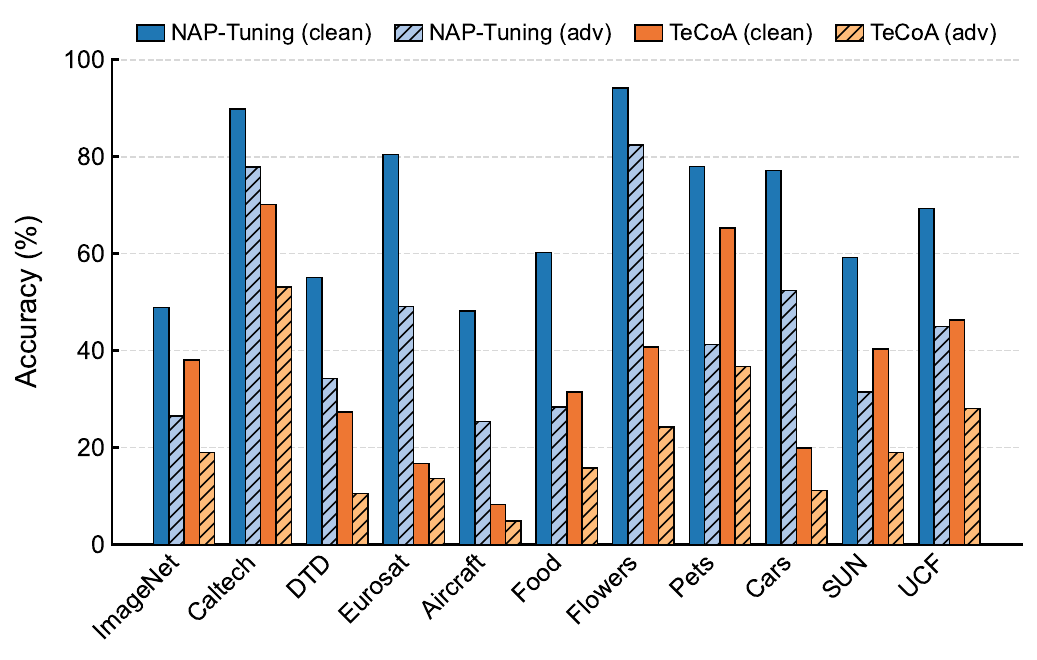} 
\caption{Performance comparison between NAP-Tuning (frozen image encoder) and TeCoA (fine-tuned image encoder).} 
\label{fig:tecoa_comparison} 
\end{figure}

\textbf{Impact of Fine-tuning.} Fig.~\ref{fig:tecoa_comparison} reveals a critical trade-off. While TeCoA achieves a smaller robustness gap (15.3\% vs. 24.2\%), it suffers a catastrophic 32.3\% drop in clean accuracy (69.1\% $\to$ 36.8\%) compared to NAP-Tuning. This failure, consistent across datasets like EuroSAT, confirms that fine-tuning the backbone disrupts the pre-trained multi-modal alignment essential for zero-shot capabilities. In contrast, NAP-Tuning preserves the encoder's generalization ability by using the Neural Augmentor for targeted feature purification, thereby maintaining strong performance on both clean and adversarial data without compromising the pre-trained representations.

\begin{table}[!t]
\caption{Training time comparison on Pets dataset. All methods trained for the same number of epochs with identical batch size on a single NVIDIA A800 GPU.}
\label{tab:training_time}
\centering
\renewcommand{\arraystretch}{1.2}
\begin{tabular}{l c c}
\toprule
\multirow{2}{*}{\textbf{Method}} & \multicolumn{2}{c}{\textbf{Training Efficiency}} \\
\cmidrule(lr){2-3}
 & \textbf{Time (minutes)} & \textbf{Relative} \\
\midrule
\multicolumn{3}{l}{\textit{Input-level Prompt Methods}} \\
AdvPT & 34 & 1.00$\times$ \\
\midrule
\multicolumn{3}{l}{\textit{Deep Prompt Methods}} \\
AdvPT-V & 91 & 2.68$\times$ \\
AdvPT-VLI & 93 & 2.74$\times$ \\
FAP & 91 & 2.68$\times$ \\
APD & 95 & 2.79$\times$ \\
NAP-Tuning (Ours) & 99 & 2.91$\times$ \\
\midrule
\multicolumn{3}{l}{\textit{Backbone Fine-tuning Methods}} \\
TeCoA & 123 & 3.62$\times$ \\
\bottomrule
\end{tabular}
\end{table}

\textbf{Computational Efficiency.} As shown in Table~\ref{tab:training_time}, NAP-Tuning maintains a favorable efficiency profile. While deep prompt methods (including ours) require $\sim2.9\times$ training time compared to the lightweight text-only AdvPT, they remain significantly more efficient than full fine-tuning. TeCoA incurs a 3.62$\times$ computational cost due to full gradient propagation through the backbone. Thus, NAP-Tuning achieves state-of-the-art robustness with a much lower computational burden than fine-tuning approaches.

\subsection{Component Ablation Studies} \label{sec:ablations}

Having demonstrated NAP-Tuning's superior performance, we now conduct a series of ablation studies to deconstruct our framework. We systematically evaluate the impact of each core design choice, from the high-level prompt architecture down to the specific configuration of our purification module.

\subsubsection{Effect of Multi-layer Prompt Architecture} \label{sec:layer}

\begin{figure*}[!t]
\centering
\includegraphics[width=\linewidth]{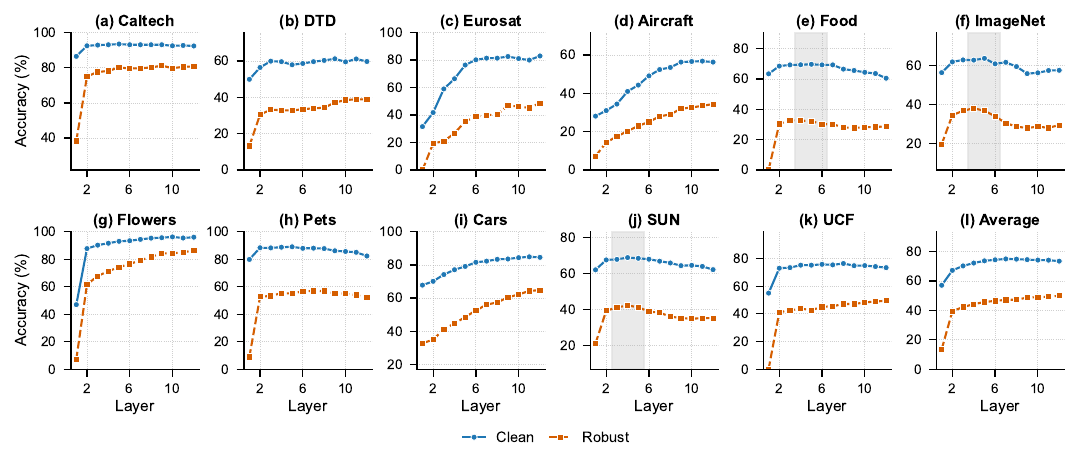}
\caption{Clean and robust accuracy across datasets when varying the number of prompt layers (1-12). Complex datasets (ImageNet, Food101, SUN397) show optimal performance at intermediate depth, while other datasets benefit from deeper architectures.}
\label{fig:layer_prompt}
\end{figure*}

To investigate how the depth of our multi-layer prompt architecture affects model robustness, we conduct an ablation study by varying the number of layers where prompts are inserted, from 1 to 12. Fig. \ref{fig:layer_prompt} presents the performance on eleven diverse datasets under both clean and adversarial conditions.

Fig.~\ref{fig:layer_prompt} confirms that increasing prompt depth generally enhances both clean and robust accuracy. However, complex datasets (e.g., ImageNet, Food101) exhibit an inverted U-shape trajectory, peaking at intermediate depths. This pattern likely stems from mild overfitting given the limited training samples (16-shot). Despite these task-specific variations, the full-depth configuration (layer=12) yields the superior average performance and consistently outperforms baselines, justifying its adoption as the default setting.

\subsubsection{Effect of the Neural Augmentor Module} \label{sec:ablation_na}

\begin{figure}[!t]
    \centering
    \begin{subfigure}
        \centering
        \includegraphics[width=0.47\linewidth]{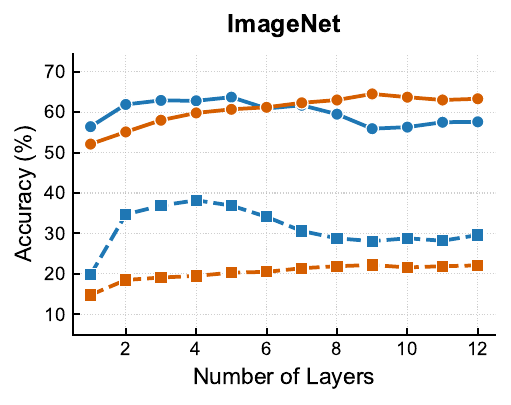}
    \end{subfigure}
    \hfill
    \begin{subfigure}
        \centering
        \includegraphics[width=0.47\linewidth]{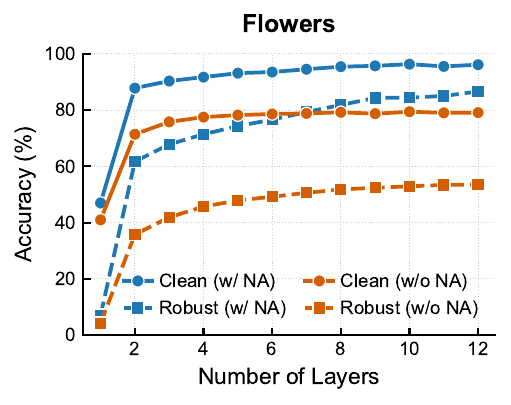}
    \end{subfigure}
    \caption{Ablation analysis comparing models with and without the Neural Augmentor (NA) across varying prompt depths. The complete model consistently outperforms the ablated variant.}
    \label{fig:na_comparison}
\end{figure}

Complementing the multi-layer analysis, Fig.~\ref{fig:na_comparison} isolates the specific contribution of the Neural Augmentor. The results unequivocally show that the complete model yields consistently higher robust accuracy across all layer configurations on both ImageNet and Oxford Flowers. A particularly interesting observation is the interaction between module capacity and task complexity. On ImageNet, the complete model peaks earlier (around Layer 4) compared to the ablated variant, which plateaus much later (around Layer 9). This shift suggests that the Neural Augmentor accelerates adaptation by providing necessary structural capacity. While this added capacity introduces a mild overfitting tendency in few-shot settings (manifesting as the inverted U-shape), the substantial performance gain confirms that the benefits of explicit feature purification far outweigh these limitations.

\subsubsection{Effect of TokenRefiner Modality}

\begin{figure}[!t]
\centering
\includegraphics[width=\columnwidth]{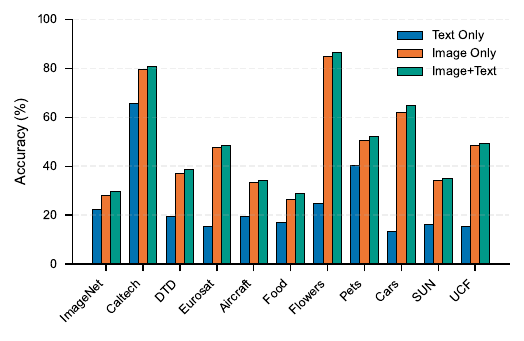} \caption{Modality-specific TokenRefiner ablation study under PGD-100 attacks. While image-side TokenRefiner provides substantial robustness benefits, the dual-modal approach consistently outperforms single-modality configurations across all datasets.}
\label{fig:modal-ablation}
\end{figure}

We analyze the modality-specific contributions in Fig.~\ref{fig:modal-ablation}. A clear hierarchy emerges: the Image-only TokenRefiner serves as the primary defense backbone, driving the vast majority of robustness gains. This aligns with the theoretical expectation that direct visual purification is essential against image-based attacks. Interestingly, the Text-only variant—despite operating on unperturbed inputs—still yields considerable robustness improvements. This suggests a mechanism of cooperative feature alignment, where the module dynamically adapts clean text embeddings to better match the purified visual features. Consequently, the full Dual-modal framework consistently outperforms single-branch configurations, confirming that text-side alignment is a critical, non-redundant complement to visual purification.

\subsubsection{Effect of TokenRefiner Architecture Depth}

\begin{figure}[!t] 
\centering 
\subfigure[ImageNet]{ 
\includegraphics[width=0.46\linewidth]{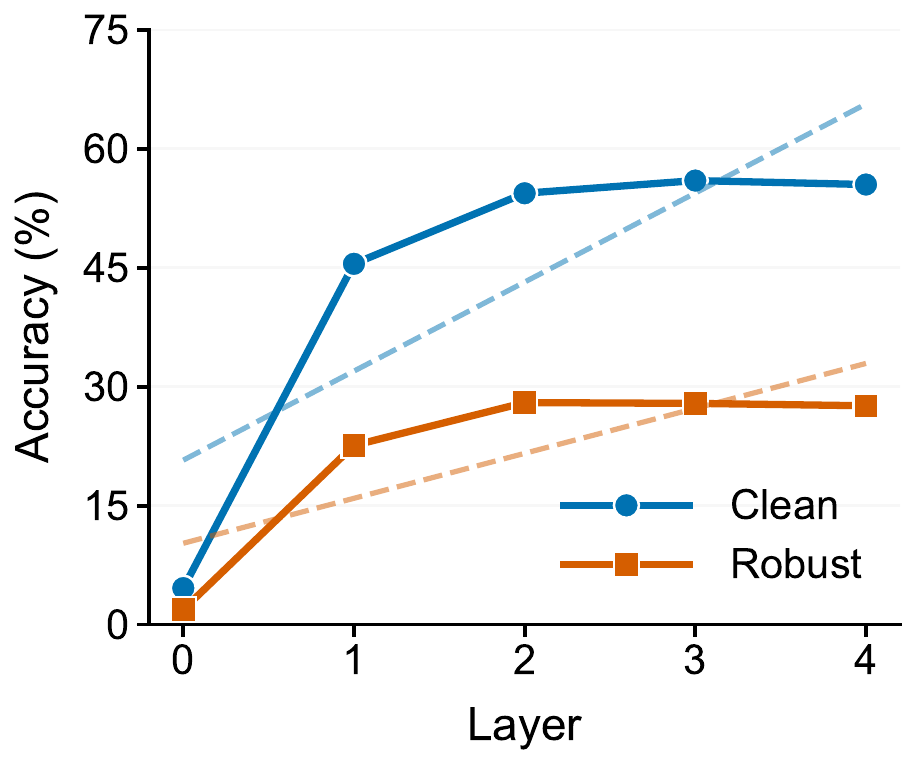} } \subfigure[Flowers]{ 
\includegraphics[width=0.46\linewidth]{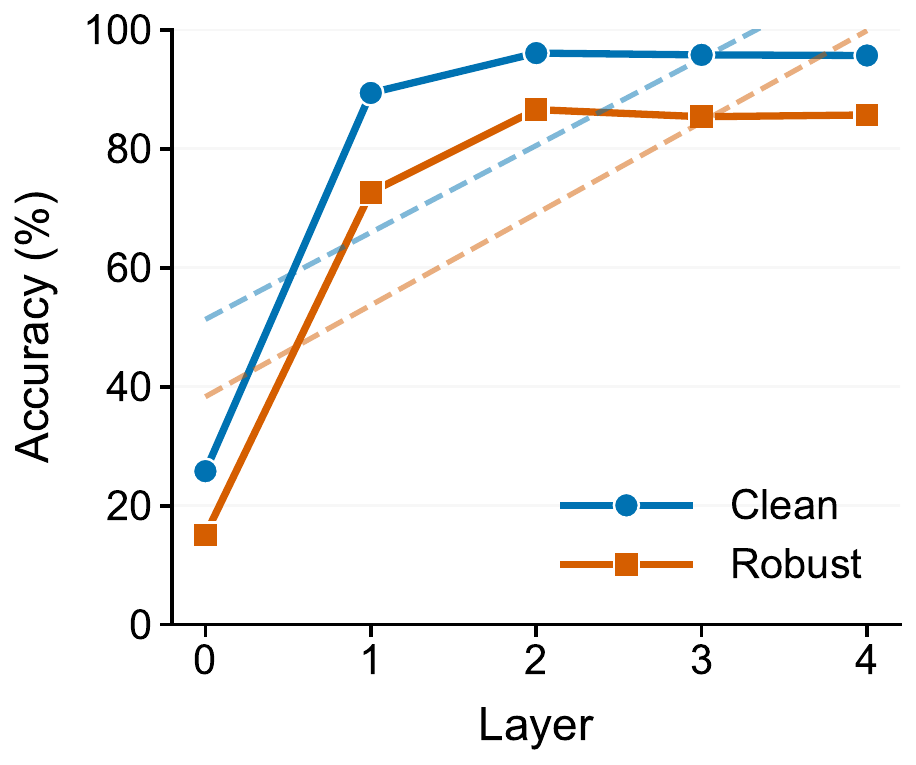} } \caption{TokenRefiner performance across different network depths. Performance improves dramatically from 0 to 2 layers before stabilizing, with consistent patterns across datasets of varying complexity.} 
\label{fig:token} 
\end{figure}

We investigate the impact of TokenRefiner network depth on model performance using Oxford Flowers (simpler task) and ImageNet (complex task) in Fig.~\ref{fig:token}. The results reveal that the TokenRefiner is not merely an enhancement but a prerequisite for optimization stability. Specifically, removing the module (layer = 0) leads to model collapse under 5-step PGD adversarial training, resulting in near-random performance on both datasets. However, increasing the depth to just 2 layers resolves this instability, yielding dramatic improvements that restore both clean and robust accuracy to competitive levels. Beyond this point, performance saturates, justifying our selection of a 2-layer architecture as the optimal configuration. Notably, this structural stability distinguishes our approach from previous methods like FAP~\cite{zhou2024few} and APD~\cite{luo2024adversarial}; whereas those methods suffer from convergence issues under strong adversarial search, our TokenRefiner enables robust optimization even under rigorous perturbation constraints. This consistent behavior across datasets of varying complexity confirms that our design generalizes well without necessitating task-specific tuning.

\subsection{In-depth Analysis} \label{sec:analysis}

Beyond ablating components, we now provide a deeper analysis of our method's core mechanism. We investigate \emph{how} NAP-Tuning works by visualizing its effect on feature space, and explore its \emph{compatibility} with other defense paradigms.

\subsubsection{Analysis of Feature Purification} \label{sec:feature_analysis}

To investigate the effectiveness of different defensive strategies in preserving feature integrity under adversarial attack, we analyze both quantitative distortion measurements and qualitative attention visualizations.

\begin{table}[t]
\centering
\caption{Feature distortion magnitude across different defensive approaches measured using Equation~\eqref{eq:delta}. Lower value is better.}
\begin{tabular}{lr}
\toprule
\textbf{Method} & \textbf{Feature Distortion ($L_2$ Norm)} \\
\midrule
Vanilla & 8.67 \\
AdvPT & 8.63 \\
APD & 5.01 \\
AdvPT-V & 4.87 \\
AdvPT-VLI & 4.76 \\
FAP & 3.58 \\
\textbf{NAP-Tuning} & \textbf{1.96} \\
\bottomrule
\end{tabular}
\label{tab:distortion}
\end{table}

\begin{table}[t]
\centering
\caption{Accuracy (\%) under $\ell_\infty$-bounded adversarial perturbations. Models are trained and evaluated with varying perturbation magnitudes (measured in $\epsilon/255$).}
\label{tab:analysis_epsilon}
\begin{tabular}{cccccc}
\toprule
\multirow{2}{*}{\textbf{Training $\boldsymbol{\epsilon}$}} & \multirow{2}{*}{\textbf{Clean Accuracy}} & \multicolumn{4}{c}{\textbf{Robust Accuracy @ Test-time $\boldsymbol{\epsilon}$}} \\
\cmidrule{3-6}
& & \textbf{1} & \textbf{2} & \textbf{4} & \textbf{8} \\
\midrule
1 & 96.1 & 86.5 & 67.0 & 17.1 & 0.1 \\
2 & 94.2 & 87.2 & 75.3 & 38.1 & 1.3 \\
4 & 87.5 & 79.2 & 75.7 & 55.9 & 11.8 \\
\bottomrule
\end{tabular}
\vspace{-0.1in}
\end{table}

\begin{figure}[!t]
    \centering
    \begin{tabular}{c@{\hspace{1mm}}c@{\hspace{4mm}}c@{\hspace{1mm}}c}
        \includegraphics[width=0.22\columnwidth]{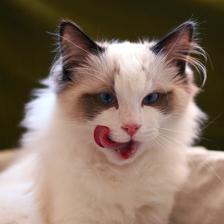} &
        \includegraphics[width=0.22\columnwidth]{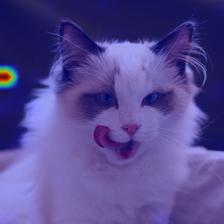} & 
        \includegraphics[width=0.22\columnwidth]{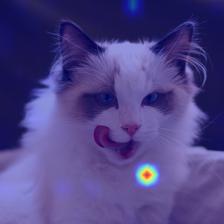} & 
        \includegraphics[width=0.22\columnwidth]{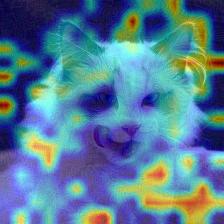} \\

        \footnotesize{Original} & \footnotesize{Adversarial} & \footnotesize{AdvPT} & \footnotesize{AdvPT-V} \\
        
        \includegraphics[width=0.22\columnwidth]{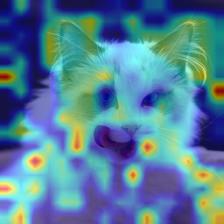} & 
        \includegraphics[width=0.22\columnwidth]{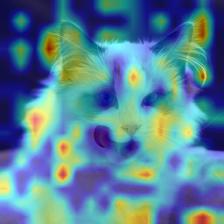} &
        \includegraphics[width=0.22\columnwidth]{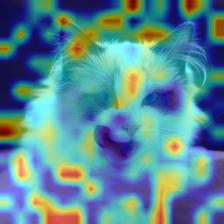} & 
        \includegraphics[width=0.22\columnwidth]{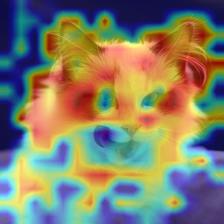} \\
        
        \footnotesize{AdvPT-VLI} & \footnotesize{APD}  & \footnotesize{FAP} &  \footnotesize{NAP-Tuning}\\
    \end{tabular}
    \caption{EigenCAM visualization of attention patterns. NAP-Tuning maintains focus on semantically meaningful regions (cat's face) even under adversarial attack, while other methods exhibit significant attention shifts toward irrelevant areas.}
    \label{fig:eigencam}
\end{figure}

Table~\ref{tab:distortion} quantifies the feature distortion magnitude under attack. A distinct capability hierarchy emerges from the results. Shallow approaches (Vanilla CLIP, AdvPT) exhibit significant distortion, confirming that input-level optimization alone fails to restrain feature corruption. Methods incorporating deeper interventions (e.g., APD, AdvPT-V) achieve moderate mitigation. In sharp contrast, NAP-Tuning outperforms all competitors by a wide margin, achieving the lowest distortion score. This represents a drastic reduction compared to the baseline, validating our method's unique ability to maintain structural integrity in the embedding space.

These quantitative findings are strongly corroborated by the EigenCAM~\cite{muhammad2020eigen} visualizations in Fig.~\ref{fig:eigencam}. While baseline models suffer from severe attention dispersion—scattering focus across irrelevant background regions—NAP-Tuning maintains precise semantic consistency. Specifically, our model retains focus on discriminative regions (e.g., facial features) even under adversarial attack, closely mirroring the attention patterns of clean inputs. This alignment between reduced feature distortion and preserved attention focus confirms our central hypothesis: adversarial vulnerability stems primarily from internal feature corruption, which the Neural Augmentor effectively rectifies.

\subsubsection{Analysis of Model-Image Defense Integration}

We investigate NAP-Tuning's capability to integrate with image-based defense mechanisms. Similar to AdvPT \cite{zhang2024adversarial}, our approach focuses exclusively on model-side adaptation, allowing it to complement image-based defense techniques. This presents an advantage over methods that already incorporate image transformations in their design, which often cannot be effectively combined with other preprocessing defenses.

\begin{figure}[!t] 
\centering 
\includegraphics[width=\linewidth]{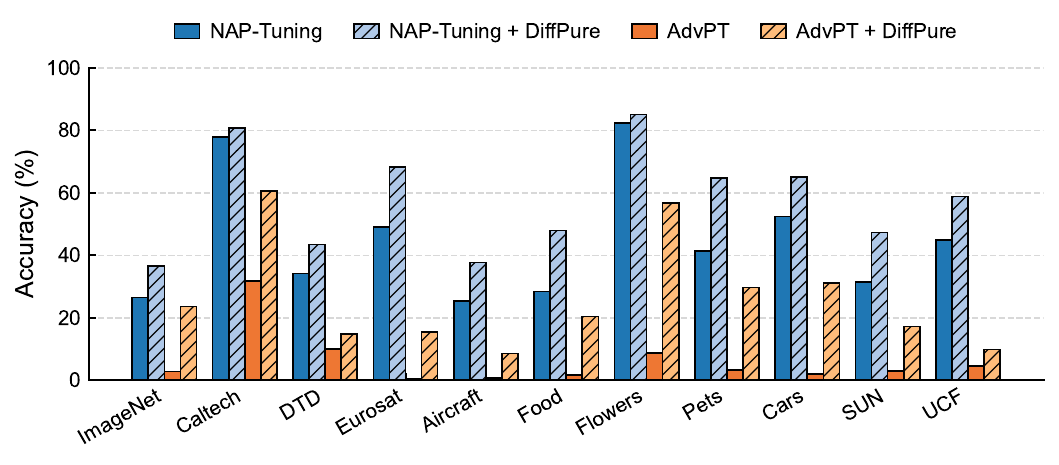} 
\caption{Performance comparison across eleven datasets demonstrates consistent improvements when combining model-centric NAP-Tuning with the image-processing DiffPure method.} 
\label{fig:pure} 
\end{figure}

\begin{figure}[!t]
\centering
\includegraphics[width=0.6\linewidth]{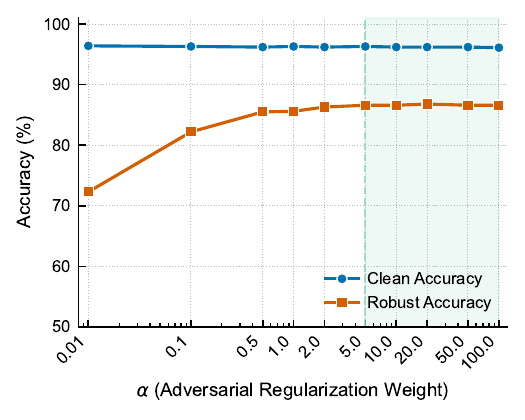}
\caption{Effect of adversarial regularization weight ($\alpha$) on clean and robust accuracy on Flowers dataset.}
\label{fig:alpha}
\end{figure}

To demonstrate this complementary capability, we evaluate NAP-Tuning combined with DiffPure \cite{nie2022diffusion}, a diffusion-based adversarial purification technique. Fig.~\ref{fig:pure} shows the combined approach consistently outperforms standalone NAP-Tuning, with an average improvement of 12.9 percentage points. These results highlight a key advantage of our model-focused approach: NAP-Tuning enhances network parameter robustness while remaining fully compatible with input-processing techniques. This orthogonality enables effective integration that addresses different aspects of adversarial vulnerability. The consistent improvements across diverse datasets suggest that combining complementary defense strategies represents a promising direction for building more robust visual recognition systems.

\subsection{Parameter Sensitivity Analysis}\label{sec:sensitivity}

Finally, we evaluate the sensitivity of NAP-Tuning to key hyperparameters and training settings. These experiments validate the robustness and stability of our approach across different configurations.

\subsubsection{Impact of Adversarial Regularization ($\alpha$)}

We investigate the sensitivity of NAP-Tuning to the adversarial regularization weight ($\alpha$) in Fig.~\ref{fig:alpha}. The analysis reveals a highly favorable decoupling of metrics: clean accuracy remains remarkably stable across the entire parameter sweep, showing negligible fluctuation regardless of the regularization strength. In contrast, robust accuracy exhibits a rapid ascent as $\alpha$ increases from minimal values, effectively stabilizing around $\alpha=2.0$. Beyond this threshold ($\alpha \geq 5.0$), performance saturates at a high level with diminishing returns. Consequently, we adopt $\alpha=5.0$ as our default configuration. This setting yields a near-optimal balance without strictly necessitating the absolute peak observed at higher values (e.g., $\alpha=20.0$). These findings confirm that our approach is insensitive to precise hyperparameter tuning, maintaining consistent effectiveness across a broad operational range.

\begin{figure*}[!t]
\centering
\includegraphics[width=\linewidth]{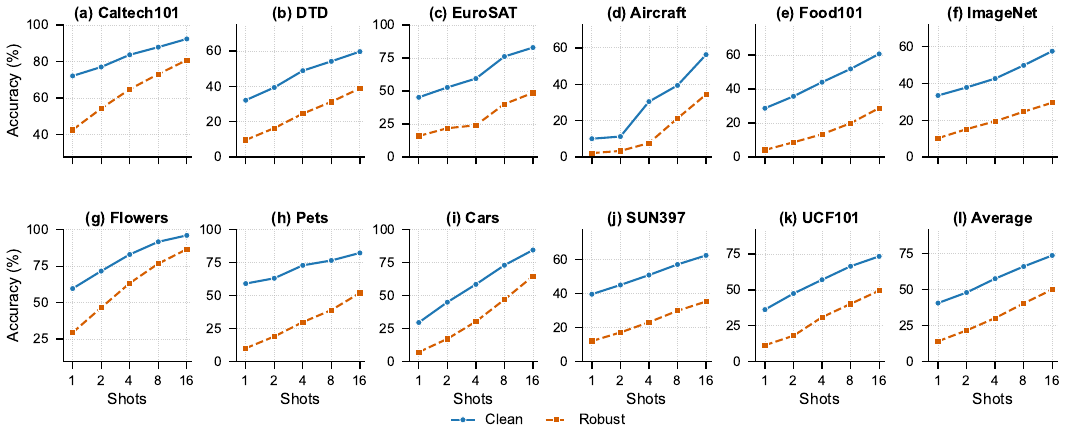}
\caption{Impact of shot count on clean and robust accuracy across eleven datasets. Each subplot (a-k) shows performance on individual datasets, while subplot (l) presents the average across all datasets.}
\label{fig:shot_analysis}
\end{figure*}

\begin{figure}[!t]
\centering
\includegraphics[width=\linewidth]{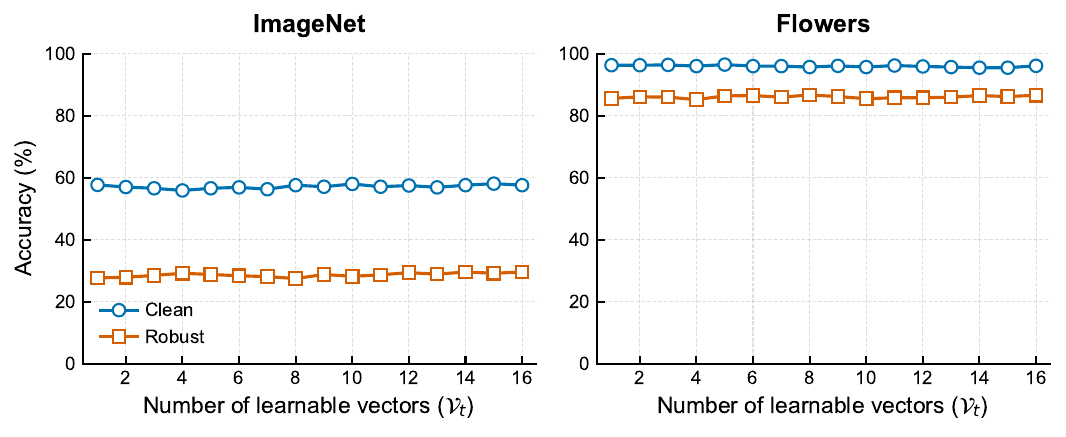}
\caption{Effect of context vector count (1-16) on ImageNet and Oxford Flowers. Performance varies minimally across different vector counts, indicating context vectors are not a primary bottleneck for model capability.}
\label{fig:context_vectors}
\end{figure}

\subsubsection{Impact of Perturbation Magnitude ($\epsilon$)}

We evaluate the interplay between perturbation budgets and model performance in Table~\ref{tab:analysis_epsilon}. A clear trade-off emerges from the data: increasing the training perturbation magnitude ($\epsilon$) substantially bolsters robustness against stronger attacks, yet this hardening incurs a discernible cost in clean accuracy. Specifically, models trained with larger budgets demonstrate vastly superior resilience when subjected to matching or higher-intensity threats. In contrast, models trained on weaker perturbations suffer from rapid degradation when facing attacks that exceed their training budget. This aligns with well-established observations in traditional adversarial training, confirming that robust prompt tuning follows the same fundamental principle: resistance to extreme perturbations is achieved by sacrificing a degree of standard generalization.

\subsubsection{Impact of Few-shot Count}

We examine the impact of shot count on model performance. Fig. \ref{fig:shot_analysis} illustrates the relationship between number of shots (1-16) and model accuracy across diverse datasets. Clean accuracy shows consistent improvement, with average accuracy increasing from 40.6\% (single shot) to 73.5\% (16 shots). More remarkably, robust accuracy exhibits steeper relative gains, rising from 14.1\% to 49.9\%.
This differential impact reveals a key insight: while few-shot learning is known to enhance clean accuracy, its benefits for robust performance are substantially more pronounced. The results indicate that robustness requires a richer representation of class concepts that becomes increasingly available with additional examples. These findings align with our observations in Section \ref{sec:layer}, confirming that increasing shot count offers tremendous potential for performance enhancement in our method, providing a straightforward path to significantly improved robustness without architectural modifications.

\subsubsection{Impact of Context Vector Count}
We investigate the sensitivity of model performance to the length of learnable context vectors ($\mathcal{V}_t$) in Fig.~\ref{fig:context_vectors}. On the complex ImageNet dataset, clean accuracy remains remarkably stable regardless of vector length, while robust accuracy exhibits only marginal gains as the number of vectors increases, eventually saturating at higher counts. This stability is even more pronounced on the simpler Flowers dataset, where both metrics maintain consistently high performance with negligible variance across all configurations. Crucially, these fluctuations are minimal when compared to the significant impact of prompt depth observed in Section~\ref{sec:layer}. This confirms that input-level representational capacity is not the primary bottleneck; rather, the structural depth of the intervention plays a far more decisive role in determining robustness than the mere quantity of input tokens.

\section{Conclusion and Discussion}

This paper presents the Neural Augmentor framework (NAP-Tuning) as a modular structural augmentation for enhancing the adversarial robustness of Vision-Language Models. By introducing internal TokenRefiners into a coordinated multi-modal and multi-layer prompting structure, we enable the explicit rectification of adversarial distortions within the feature manifold. Our empirical results across 11 benchmarks demonstrate that this internal purification mechanism significantly outperforms conventional input-side alignment methods, particularly under rigorous evaluation protocols. Crucially, the plug-and-play nature of the Neural Augmentor ensures high compatibility with frozen VLM backbones, providing a scalable and parameter-efficient defense for robust multi-modal learning.

Our analysis underscores that adversarial vulnerability in VLMs stems primarily from internal feature distortions that input-side prompt tuning alone cannot fully rectify. By shifting the defense focus to structural feature-level intervention, NAP-Tuning provides a principled way to restore representational integrity. Future research will focus on developing adaptive TokenRefiners capable of dynamic adjustment based on perturbation intensity, as well as extending this purification logic to generative multimodal architectures beyond contrastive learning paradigms.

\bibliographystyle{IEEEtran}
\bibliography{main}



\begin{IEEEbiography}[{\includegraphics[width=1in,height=1.25in,clip,keepaspectratio]{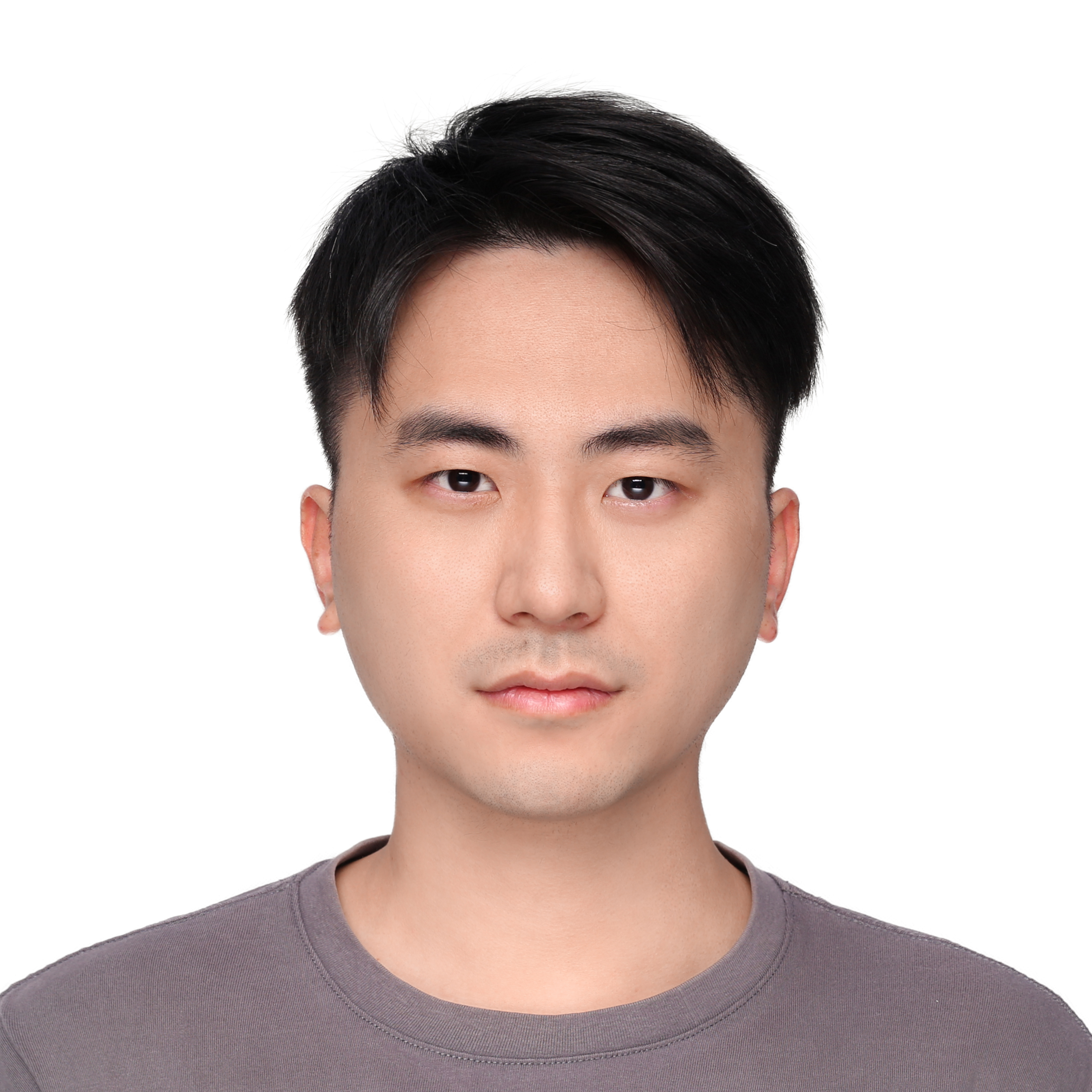}}]{Jiaming Zhang}
received the Ph.D. degree in computer science from Beijing Jiaotong University, China. He was a visiting researcher at Fudan University and a postdoctoral researcher at The Hong Kong University of Science and Technology. His research interests include trustworthy artificial intelligence and multimedia computing.
\end{IEEEbiography}

\begin{IEEEbiography}[{\includegraphics[width=1in,height=1.25in,clip,keepaspectratio]{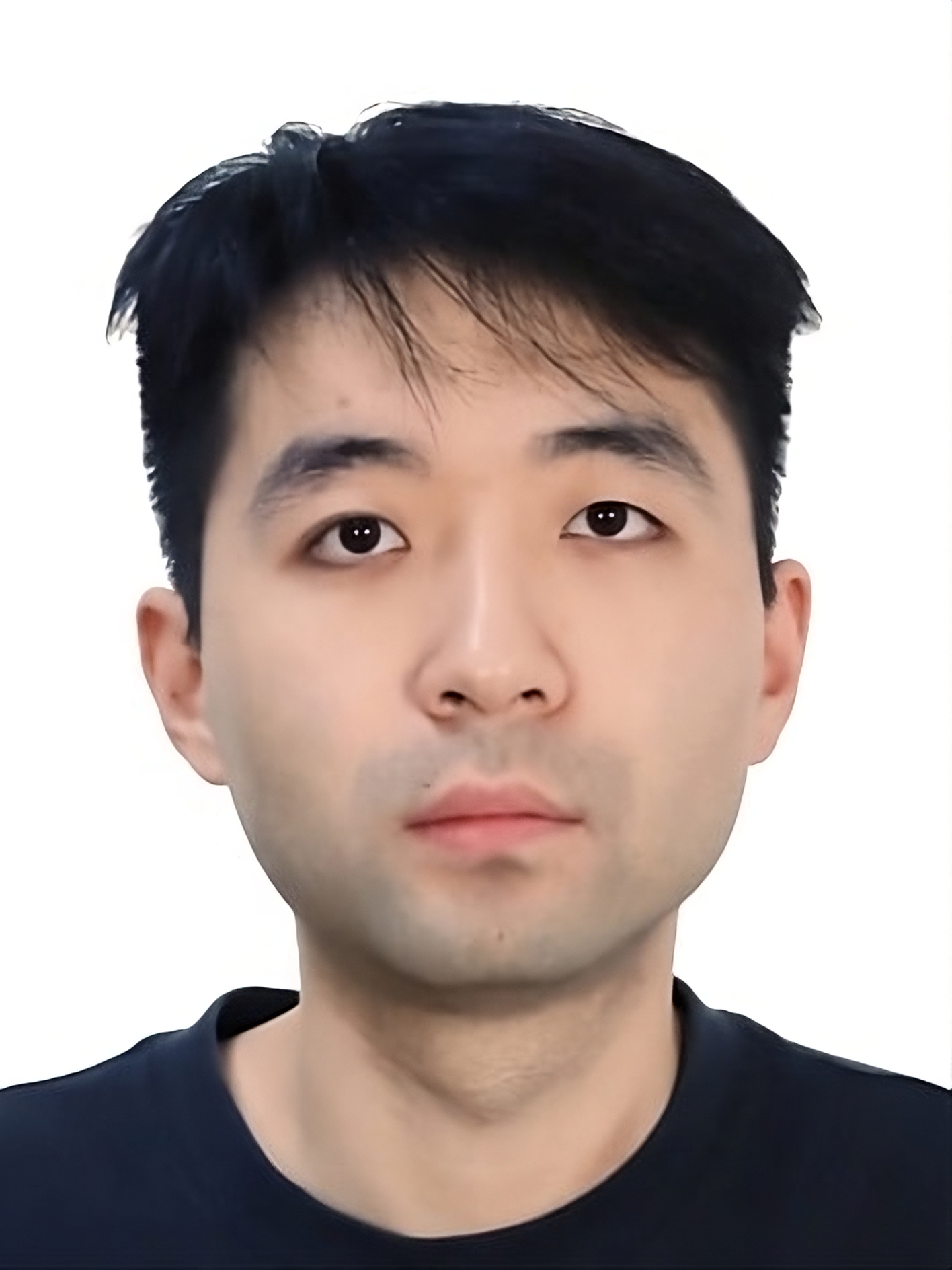}}]{Xin Wang}
received the Master's degree in the Faculty of Artificial Intelligence in Education, Central China Normal University, Wuhan, China, in 2021. He is currently working toward the Ph.D. degree in Computer Science with the School of Computer Science, Fudan University, Shanghai, China. His research interests mainly include computer vision and trustworthy machine learning.
\end{IEEEbiography}

\begin{IEEEbiography}[{\vspace{-20pt}\includegraphics[width=1in,height=1.25in,clip,keepaspectratio]{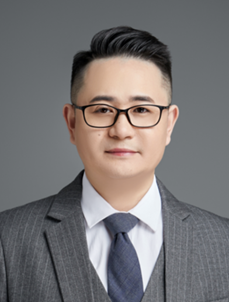}}]{Xingjun Ma}
is currently an associate professor at Fudan University. He received his PhD degree from The University of Melbourne. His main research area is trustworthy AI, aiming to design secure, robust, explainable, privacy-preserving, and fair machine learning models for real-world AI applications. He has published at top-tier conferences and journals, including ICLR, NeurIPS, and ICML. He also serves as an area chair for a number of AI conferences. 
\end{IEEEbiography}

\begin{IEEEbiography}[{\includegraphics[width=1in,height=1.25in,clip,keepaspectratio]{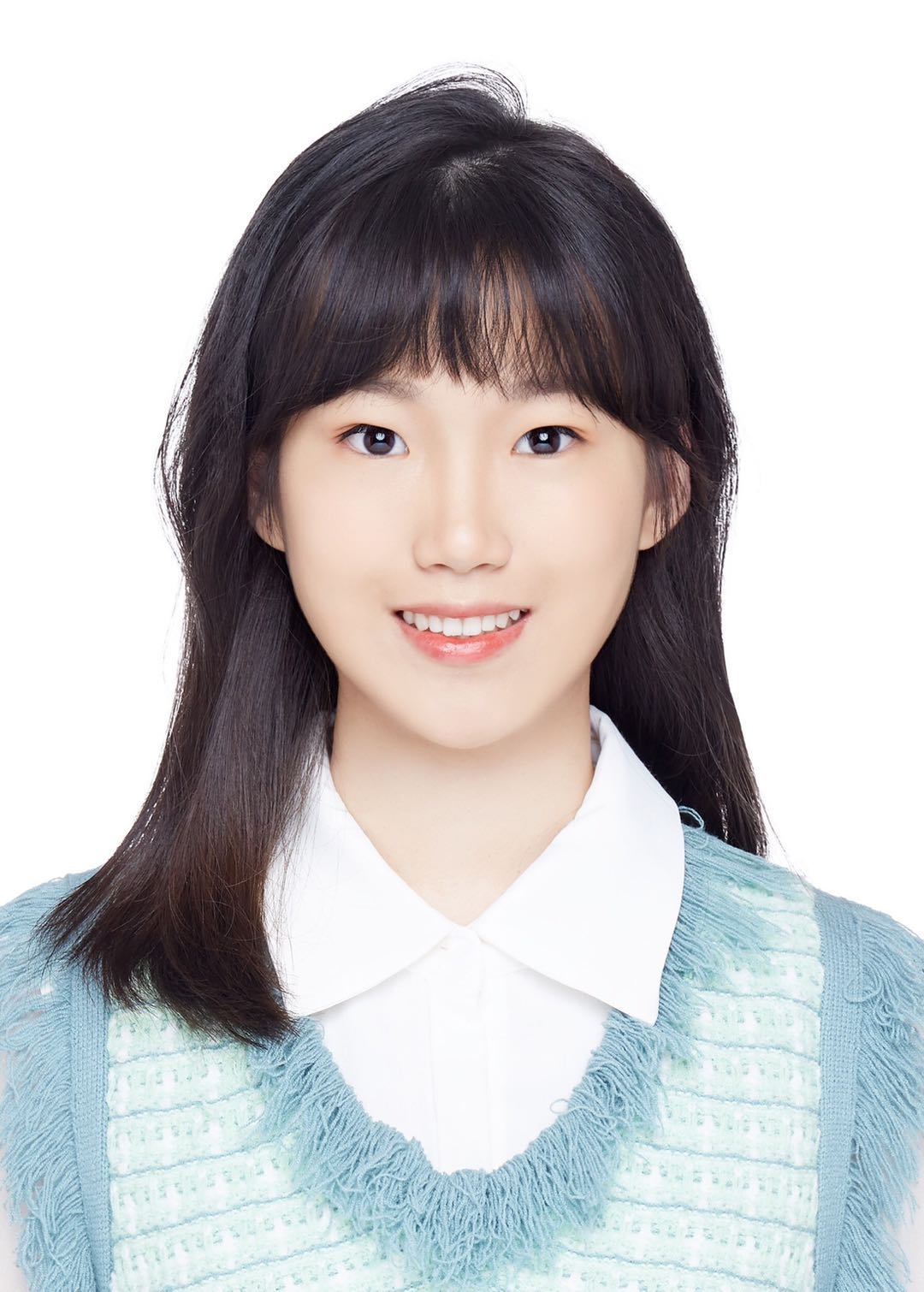}}]{Lingyu Qiu}
is a Marie Skłodowska-Curie (MSCA) PhD student in Computer Science at The University of Naples Federico II, Italy. She received her master degree from Nanjing University of Aeronautics and Astronautics, China. Her research lies in trustworthy machine learning and federated learning that ensure sustainability and trustworthiness in various sectors.
\end{IEEEbiography}

\begin{IEEEbiography}[{\includegraphics[width=1in,height=1.25in,clip,keepaspectratio]{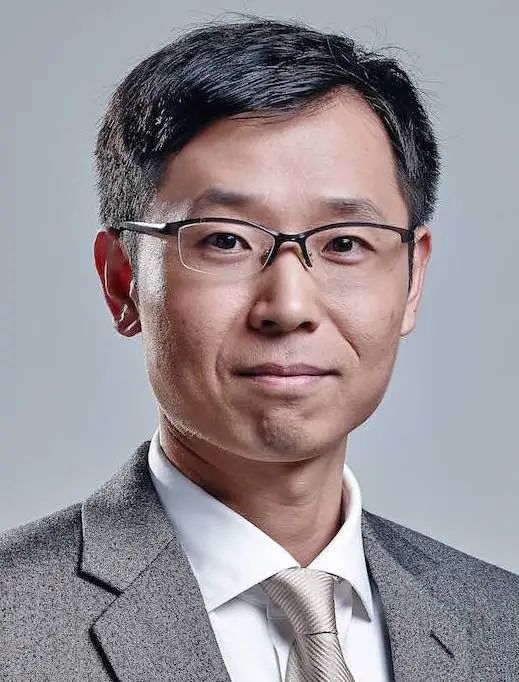}}]{Yu-Gang Jiang}
received the Ph.D. degree in Computer Science from the City University of Hong Kong, in 2009. He worked as a postdoctoral research scientist with Columbia University, New York, from 2009 to 2011. He is currently a vice president and Chang Jiang scholar distinguished professor of computer science with Fudan University, Shanghai, China. His research lies in the areas of multimedia, computer vision, and trustworthy AGI. He is a fellow of the IEEE and the IAPR.
\end{IEEEbiography}

\begin{IEEEbiography}[{\includegraphics[width=1in,height=1.25in,clip,keepaspectratio]{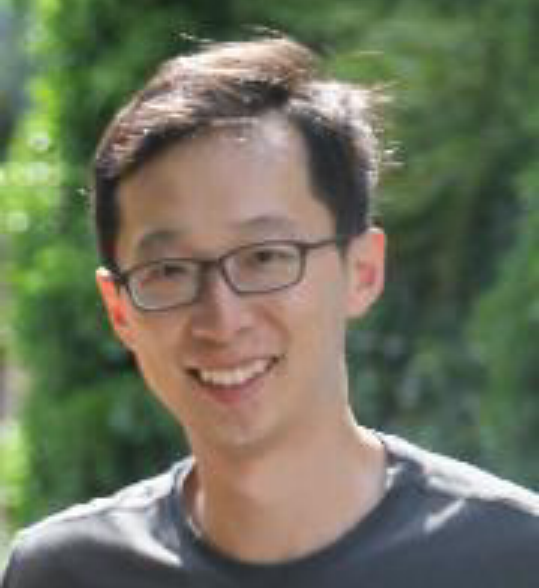}}]{Jitao Sang}
received the Ph.D. degree  from CASIA, Beijing, China with the Special Prize of President Scholarship. He is a Professor with Beijing Jiaotong University, China. His research interests include multimedia content analysis, trustworthy machine learning, AI alignment, and AI Agent.
\end{IEEEbiography}

\end{document}